\documentclass[letterpaper]{article}
\usepackage{aaai24}

\usepackage{times}  
\usepackage{helvet}  
\usepackage{courier}  
\usepackage[hyphens]{url}  
\usepackage{graphicx} 
\usepackage{natbib}  
\usepackage{caption} 
\usepackage{url}            
\usepackage{booktabs}       
\usepackage{microtype}      
\usepackage{xcolor}         
\usepackage{CJK}
\usepackage{wrapfig}
\usepackage{footnote}
\makesavenoteenv{table}
\usepackage{color}
\usepackage{algorithm}
\usepackage{algorithmic}

\usepackage{amssymb}
\usepackage{pifont}
\newcommand{\cmark}{\ding{51}}%
\newcommand{\xmark}{\ding{55}}%

\newcommand{\black}[1]{{\color{black}{#1}}}
\newcommand{\blue}[1]{{\color{blue}{#1}}}
\usepackage{hhline}
\usepackage{supertabular}
\usepackage{multirow}

\usepackage{subcaption}
\usepackage{pgfplots}
\usepgfplotslibrary{colorbrewer}
\pgfplotsset{compat = 1.15, cycle list/Set1-8} 
\usetikzlibrary{pgfplots.statistics, pgfplots.colorbrewer} 
\usepackage{pgfplotstable}
\usepackage{filecontents}
\usepackage{longtable}
\usepackage{newfloat}
\usepackage{listings}
\DeclareCaptionStyle{ruled}{labelfont=normalfont,labelsep=colon,strut=off} 
\floatstyle{ruled}
\newfloat{listing}{tb}{lst}{}
\floatname{listing}{Listing}

\usepackage{color}

\usepackage{colortbl}
\definecolor{mygray}{gray}{0.8}
\newcommand{\myred}{\black}

\title{Xiezhi: An Ever-Updating Benchmark for Holistic Domain Knowledge Evaluation}
\author{
    Zhouhong Gu\textsuperscript{\rm 1}\thanks{Equal Contribution},
    Xiaoxuan Zhu\textsuperscript{\rm 1\ *},
    Haoning Ye\textsuperscript{\rm 1},
    Lin Zhang\textsuperscript{\rm 1},
    Jianchen Wang\textsuperscript{\rm 1},
    Yixin Zhu\textsuperscript{\rm 1},\\
    Sihang Jiang\textsuperscript{\rm 1},
    Zhuozhi Xiong\textsuperscript{\rm 1},
    Zihan Li\textsuperscript{\rm 1},
    Weijie Wu\textsuperscript{\rm 1},
    Qianyu He\textsuperscript{\rm 1},
    Rui Xu\textsuperscript{\rm 1},
    Wenhao Huang\textsuperscript{\rm 1},
    Jingping Liu\textsuperscript{\rm 2},
    Zili Wang,
    Shusen Wang,
    Weiguo Zheng\textsuperscript{\rm 3},
    Hongwei Feng\textsuperscript{\rm 1}\thanks{Corresponding Authors},
    Yanghua Xiao\textsuperscript{\rm 1,4\ $\dagger$}\thanks{Yanghua Xiao is also a member of Research Group of Computational and AI Communication at Institute for Global Communications and Integrated Media, Fudan University.},
}
\affiliations{
    \textsuperscript{\rm 1}Shanghai Key Laboratory of Data Science, School of Computer Science, Fudan University, China\\
    \textsuperscript{\rm 2}School of Information Science and Engineering, East China University of Science and Technology\\
    \textsuperscript{\rm 3}School of Data Science, Fudan University\\
    \textsuperscript{\rm 4}Fudan-Aishu Cognitive Intelligence Joint Research Center\\
    \{zhgu22, xxzhu22\}@m.fudan.edu.cn, \{hwfeng, shawyh\}@fudan.edu.cn
}

\begin{document}
\begin{CJK}{UTF8}{gbsn}

\maketitle

\begin{abstract}

New Natural Langauge Process~(NLP) benchmarks are urgently needed to align with the rapid development of large language models (LLMs). 
We present Xiezhi, the most comprehensive evaluation suite designed to assess holistic domain knowledge.
Xiezhi comprises multiple-choice questions across 516 diverse disciplines ranging from 13 different subjects with 249,587 questions and accompanied by \myred{Xiezhi-Specialty with 14,041 questions and Xiezhi-Interdiscipline with 10,746 questions}. 
We conduct evaluation of the 47 cutting-edge LLMs on Xiezhi.
Results indicate that LLMs exceed average performance of humans in science, engineering, agronomy, medicine, and art, but fall short in economics, jurisprudence, pedagogy, literature, history, and management.
All the evaluation code and data are open sourced in \url{https://github.com/MikeGu721/XiezhiBenchmark}

\end{abstract}

\section{Introduction}
Domain knowledge encompasses an in-depth comprehension of the world, necessitating the cultivation of various cognitive skills, such as memorization, abstraction, logical thinking, reasoning, and imagination. 
Human has exhibited unparalleled proficiency in domain knowledge, far exceeding any machine learning models in a long time.
Nevertheless, recent advancements in Large Language Models (LLMs), including Bloom~\cite{scao2022bloom}, Llama~\cite{touvron2023llama}, ChatGLM~\cite{du2022glm}, GPT4~\cite{openai2023gpt4, bubeck2023sparks} and so many other models, have shown remarkable capabilities in domain text understanding~\cite{wei2022emergent}.
It is time to propose more comprehensive and more prospective evaluations than before to explore whether LLMs have actually acquired knowledge, or just acquired a better imitation ability~\cite{srivastava2022bigbench}.

Constructing benchmarks is crucial for automatic evaluation as benchmarks facilitate efficient, systematic, and scalable comparisons among models. 
However, as LLMs continue to grow in size and complexity, they exhibit outstanding performance across a wide range of domain-specific tasks.
This makes even the newly released benchmarks like MMLU~\cite{hendrycks2021measuring}, BIG-bench~\cite{srivastava2022bigbench} or HELM~\cite{liang2022holistic} all lag behind the capabilities of the LLMs quickily~\cite{suzgun2022big-bench-hard}.

Considering LLMs' performance, we conclude that the benchmark used to evaluate LLMs should meet the following needs:
(1) \textbf{Needs to cover more tasks~\cite{srivastava2022bigbench}}: 
Cutting-edge LLMs have integrated multiple capabilities into unified Text-to-Text transformer models~\cite{raffel2020t5}.
Therefore, the evaluation of LLMs should focus on abilities in multiple tasks. 
(2) \textbf{Needs to manifest the disparities among LLMs~\cite{huang2023ceval}}: 
Considering the emergent capacity of the models~\cite{wei2022emergent}, it is likely that the SoTA LLMs \myred{by learning knowledge in different domains}, now have a certain level of performance in all domains.
To accurately evaluate the distinctions of LLMs with varying capacities, the benchmark should consider breaking down the evaluation dimensions into more detailed categories. 
This will allow for a more precise assessment of each model's capabilities and provide valuable insights into their relative strengths and weaknesses.
(3) \textbf{Needs to go ahead of the training set~\cite{bubeck2023sparks}}: 
As LLMs are trained on increasingly extensive corpora, newly released benchmarks may become part of the LLMs' training data much sooner than before.
A prerequisite for effective evaluation is to ensure that the benchmarks are fresher than the training data used by LLMs.

\begin{figure*}[t]
\begin{center}
\resizebox{\textwidth}{!}{
\includegraphics{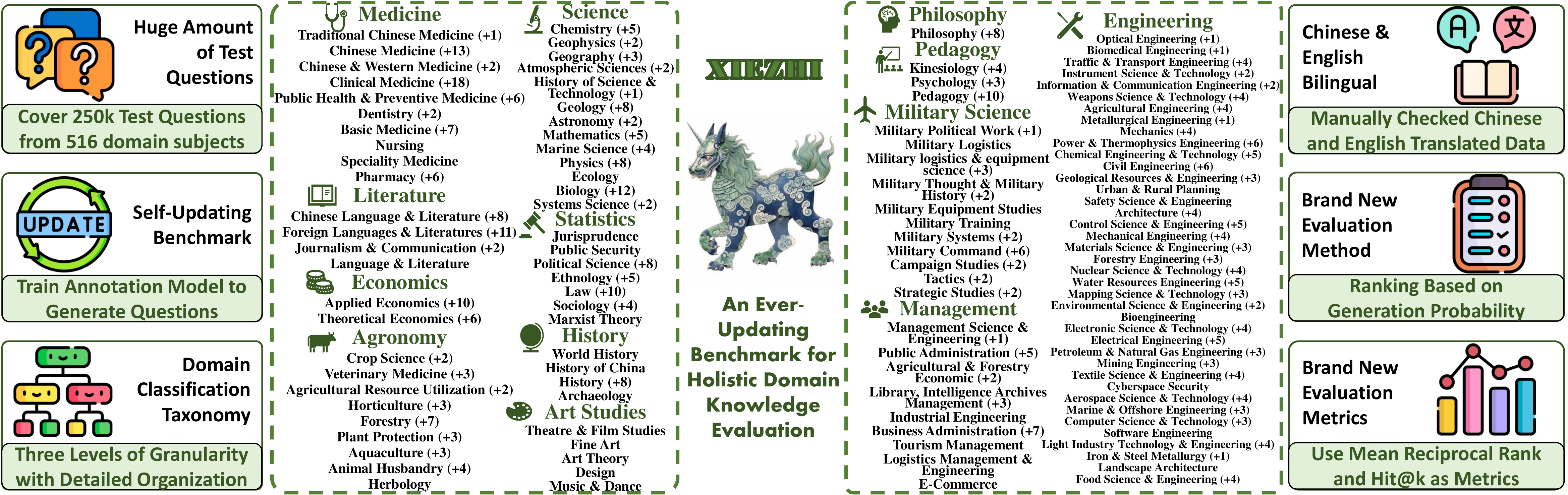}
}
\end{center}
\caption{
In Chinese mythology, the Xiezhi is a legendary creature known for its ability to discern right from wrong and uphold justice. 
Xiezhi Benchmark encompasses 13 distinct disciplinary categories, 118 sub-disciplines, and 385 further fine-grained disciplines, aiming to provide an extensive domain taxonomy and benchmark for fair, effective, and comprehensive domain evaluation.
\myred{The number adjacent to the first-level discipline signifies the number of second-level disciplines that are further divided in Chinese discipline taxonomy.}
}
\label{fig:subjectclassification}
\end{figure*}

In light of the aforementioned needs, we propose a comprehensive, multi-disciplinary, auto-updating benchmark for domain knowledge evaluation.
We call this benchmark Xiezhi, named after a mythical creature that symbolizes fairness and judgement.
Xiezhi consists of 249587 questions with 516 disciplines, ranging from 13 different categories: philosophy, economics, law, education, literature, history, natural sciences, engineering, agriculture, medicine, military science, management, and arts.
These 516 disciplines are derived from the Chinese Disciplinary Taxonomy, a comprehensive hierarchical classification system of domain knowledge proposed by the Chinese Ministry of Education and widely acknowledged in China. 
We manually selected and annotated 20,124 questions from the Chinese Graduate Entrance Examination covering these 516 labels to form the Xiezhi-Meta dataset.
Xiezhi-Meta is used to train an annotation model capable of estimating the relevance between questions and disciplinary labels.
\myred{
The annotation model subsequently tag disciplinary labels to 170k multiple-choice questions originating from diverse examinations, along with 80k multiple-choice questions auto-generated from academic surveys.}
To facilitate the usage of Xiezhi and align with the inclination that ``consolidate increasing capabilities into single LLMs'', we also present Xiezhi-Specialty and Xiezhi-Interdiscipline \myred{in both Chinese and English verision}, consisting of \myred{14,041 and 10,746 respectively} more balanced, less sensitive, and less China-centric questions. 
Xiezhi-Specialty encompasses questions solvable using knowledge from a single domain, while Xiezhi-Interdiscipline incorporates questions necessitating knowledge from multiple domains for resolution.

To give more precise evaluation results, we propose a new evaluation setting in this paper.
We set 50 options for each multiple-choice question, as previous researchers use only 4 options, resulting in significantly reducing the accuracy of random guessing and thus better revealing the model's real capabilities. 
We rank all options of each model in generation probability, as previous researchers use instructions to query the choice made by each model, to avoid inaccurate evaluations due to model's inability in answering multiple-choice questions or errors in the generated content extraction.

To provide a detailed analysis of current development status of LLMs,
as well as to demonstrate the effectiveness of the Xiezhi Benchmark and our proposed evaluation process,
we conduct experiments on \myred{47 famous LLMs} across four benchmarks proposed in different works in our evaluation setting. 
The experiments \myred{are conducted under in 0-shot, 1-shot, 3-shot demonstration setting, which is using small number of examples to demonstrate how to solve a question}, with all LLMs being evaluated on both Chinese and English versions of Xiezhi. 
This enables us to analyze the LLM results based on their optimal performance.
Results show that the best-performing LLMs\myred{, when tested via multiple-choice questions,}  have surpassed the level of average practitioners in science, engineering, agronomy, and medicine in multiple-choice form of .
But humans still greatly outperform all LLMs in domains of economics, jurisprudence, pedagogy, literature, history, and management.
We also examined the differences in performance of various LLMs across different benchmarks.
\myred{
Compared to existing knowledge evaluation benchmarks, Xiezhi covers the broadest range of domains, incorporates the highest quantity of questions, and consists of the most current data.
As shown in our experiments, due to the vast diversity of knowledge domains covered in Xiezhi and its fifty-to-one evaluation method, even marginal improvements in any aspect of a model can be accurately assessed.
As such, it is most proficient in discerning the capability differences among various LMs, spanning from GPT-4 to LLMs with only 560M parameters. Consequently, it serves as the most appropriate benchmark for evaluating LLMs of differing competencies.
}

\section{Related Works}

\subsection{Large Language Models}
Recently, various companies released their LLMs, such as BARD, ERNIE Bot, Bloom~\cite{scao2022bloom}, pythia~\cite{biderman2023pythia}, Llama~\cite{touvron2023llama}, Claude, ChatGPT~\cite{openaichatgpt}, GPT-4~\cite{openai2023gpt4}, and ChatGLM~\cite{du2022glm}.
Apart from their outstanding performance on trained tasks, researchers have also discovered that they emerge to have strong performance on many unseen tasks~\cite{zhou2023comprehensive, chung2022scaling}.
Consequently, the evaluation of LLMs' capabilities should focus more on a wide range of tasks over numerous diverse domains and contain samples with different 
difficulty levels.

The development of LLMs has spurred the growth of a series of small-scale conversational LLMs, such as Alpaca~\cite{alpaca}, Vicuna~\cite{vicuna2023}, H2Ogpt~\cite{2023h2ogpt}, and Moss~\cite{sun2023moss}.
Most of these small conversational LLMs are fine-tuned based on existing pre-trained LLMs through high-quality dialog data generated from LLMs~\cite{ji2023belle, xu2023baize} by parameter-efficient tuning methods~\cite{hu2021lora, hu2023LLM}.
In order to achieve excellent performance, these models continuously acquire the latest data from the internet, and their iteration speed is much faster than LLMs.
Any new benchmark will quickly become outdated as it is incorporated into the model's training data.

\subsection{Benchmarks for Knowledge Evaluation}

A number of studies concentrate on assessing a model's knowledge and reasoning ability. 
Certain works, including HellaSwag~\cite{zellers2019hellaswag}, Physical IQA~\cite{bisk2020piqa}, and CosmosQA~\cite{huang2019cosmos}, focus on evaluating the understanding of LLMs' commonsense knowledge. 
Meanwhile, other research, such as MMLU~\cite{hendrycks2021measuring}, AGI-Eval~\cite{zhong2023agieval}, MMCU~\cite{zeng2023measuring}, C-Eval~\cite{huang2023ceval}, M3KE~\cite{liu2023m3ke}, LexTreme~\cite{niklaus2023lextreme}, \myred{Big-Bench~\cite{srivastava2022bigbench} and BIG-Bench-Hard~\cite{suzgun2022big-bench-hard}} target at evaluating the models' proficiency in domain knowledge.
However, whether these benchmarks provide effective evaluations for all language models remains debatable. 
This is because only LLMs with super abilities show disparities on their datasets, while small LLMs only perform at a level close to random guessing, leading to different evaluation researches having different or even contradictory results on small LLMs~\cite{huang2023ceval, li2023cmmlu}.
Furthermore, as the training corpora for models become increasingly larger, these benchmarks might lose their evaluative significance shortly after they are proposed, due to their incorporation into the training sets of LLMs.

Moreover, the rise of the generative LLMs presents its own difficulties in evaluation~\cite{sai2022nlg-metric}. 
Beginning with MMLU~\cite{hendrycks2021measuring}, numerous works have proposed to use of multiple-choice questions to assess generative models. 
Recently, a variety of evaluation studies, such as SuperClue\myred{~\footnote{https://github.com/CLUEbenchmark/SuperCLUE}}, employed an identical prompt to query all LLMs and do extraction to obtain the choice made by these LLMs.
This approach requires models to have strong abilities in instruction understanding especially in multiple-choice answering, as many LLMs are unable to meet that needs, leading to unfair evaluation results.

\section{Xiezhi Benchmark}
\label{sec:benchmark}
\subsection{Chinese Discipline Taxonomy}
\label{031}
Chinese Discipline Taxonomy, developed by the Chinese Ministry of Education, organizes disciplines of different domains in college education.
The taxonomy divides all domains into different disciplines categories and various levels of disciplines.
The meanings of these levels are as follows:

\textbf{Discipline Categories}: 
This is the highest level of discipline taxonomy, divided according to the nature, characteristics of subjects. 
There are 14 subject categories in Chinese Discipline Taxonomy, including philosophy, economics, law, education, literature, history, science, engineering, agriculture, medicine, military science, management, art, and Inter-discipline.

\textbf{First-level disciplines}: 
A discipline category is divided into numerous first-level disciplines, each possessing relatively independent research content.
For example, the ``Economics'' category is divided into first-level disciplines ``Applied Economics'' and ``Theoretical Economics'', and ``Art Studies'' consist of ``Theatre \& File Studies'', ``Fine Art'' and so on.

\textbf{Second-level disciplines}: These disciplines represent more subdivided areas of study or topics within the first-level discipline.
For example, within the first-level discipline of ``Applied Economics'', further divisions include ``Financial Markets'', ``Banking'', ``Insurance'' and many other second-level disciplines.

As shown in Fig.~\ref{fig:subjectclassification}, Xiezhi Benchmark consists of a total of 13 disciplinary categories, 118 first-level disciplines, and 385 second-level disciplines as question labels.
The detailed information on the disciplines and the question amount used in Xiezhi Benchmark is listed in Tab.\myred{~\ref{table_data}} in Appendix.

\subsection{Dataset Construction} \label{032}
\subsubsection{Data collection}
Xiezhi consists of 249,587 questions from mainly two different sources. 
The first category includes nearly 170k multiple-choice questions collected from six different examinations in China: \textit{elementary school exams}, \textit{middle school entrance exams}, \textit{college entrance exams}, \textit{undergraduate exams}, \textit{graduate entrance exams}, and \textit{adult education exams}.
\myred{These questions are all open sourced and many Chinese knowledge evaluation dataset have employed these questions~\cite{huang2023ceval,liu2023m3ke}.
The second category comprises of nearly 80k multiple choice questions generated from Chinese open-source academic surveys or reviews, which is a result come from our auto updating method.
}


\begin{figure*}[t]
\begin{center}

\resizebox{\textwidth}{!}{
\includegraphics{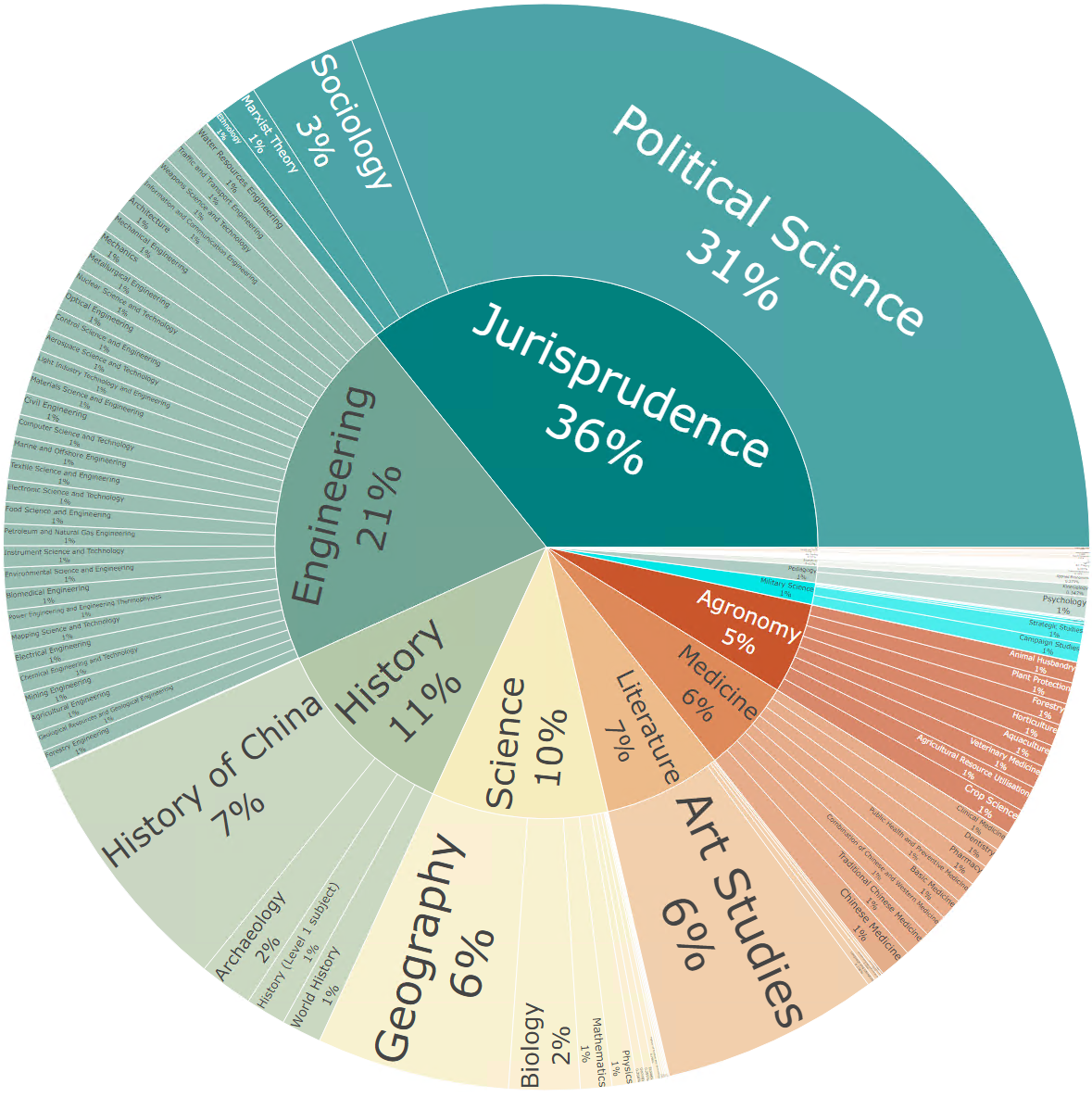}
\includegraphics{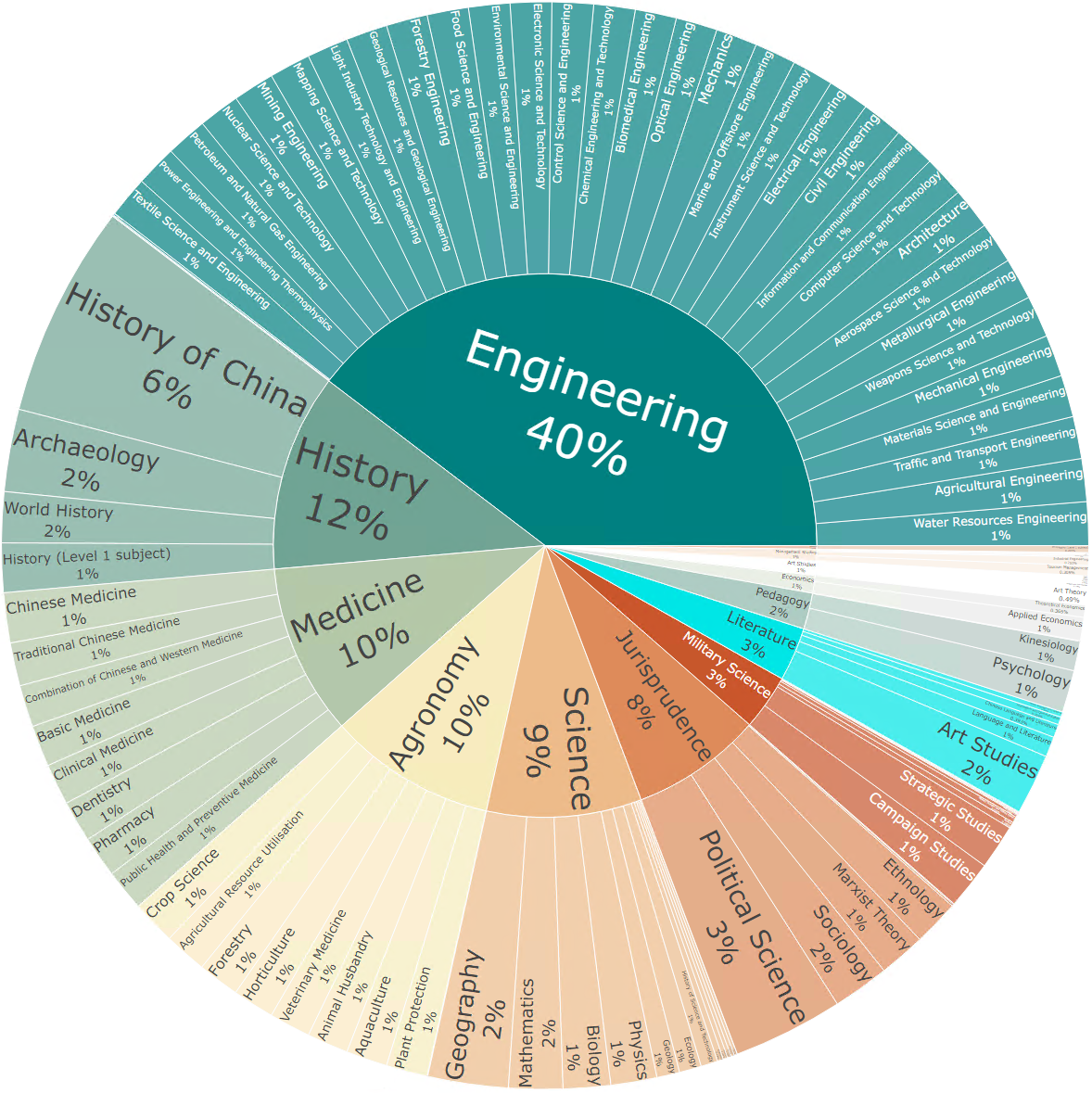}
\includegraphics{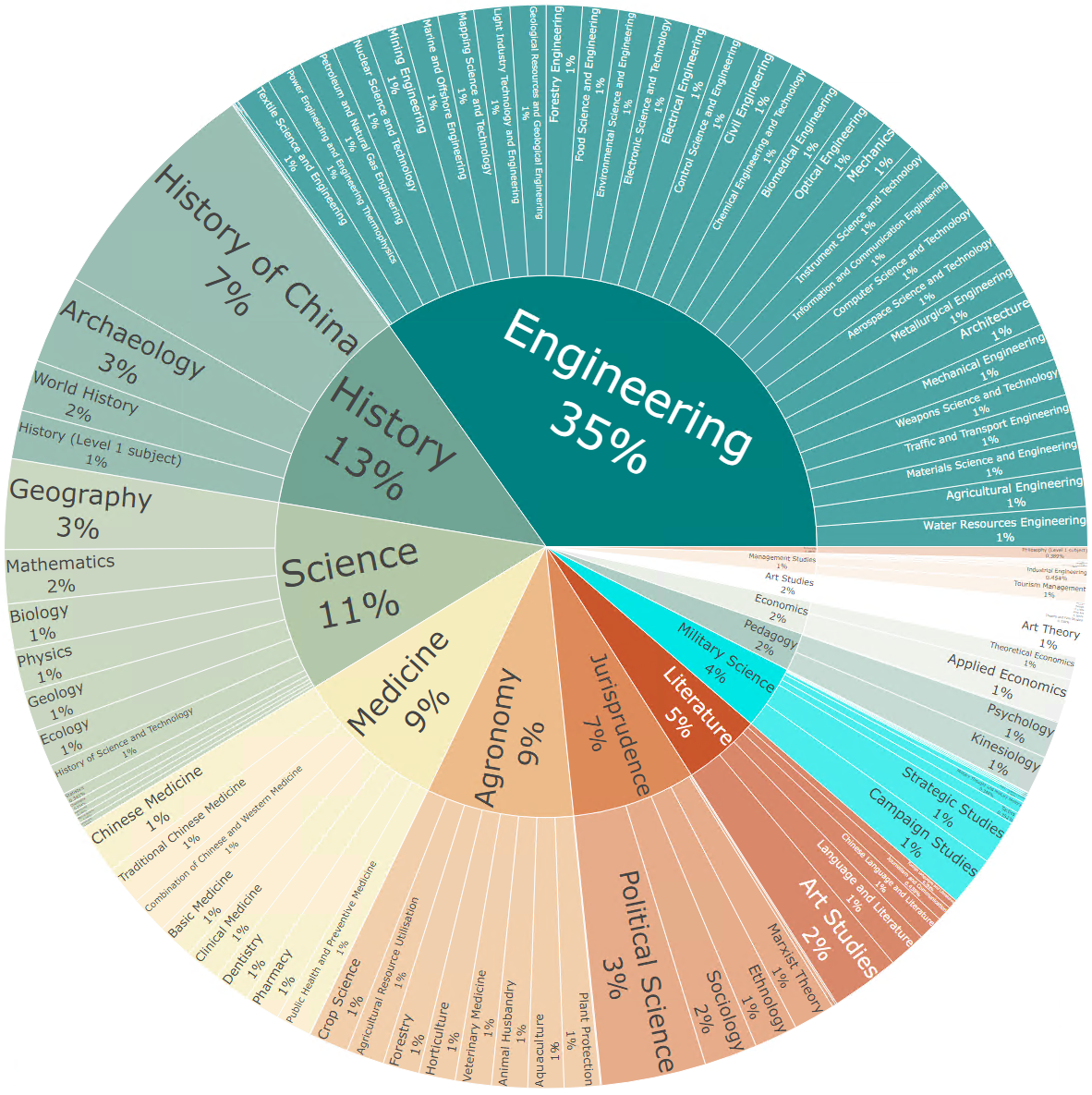}
}
\end{center}
\caption{
The figure on the right is the statistics of all questions collected by Xiezhi.
The middle figure shows statistics for Xiezhi-Specialty and the left shows Xiezhi-Interdiscipline.}
\label{fig:benchstatistic}
\end{figure*}

\subsubsection{Auto Updating}
\label{autoupdating}
Our auto-updating method comprises three primary components: 
the construction of Xiezhi-Meta dataset,
the generation of questions from open academic documents,
and the automated annotation process.

\myred{
\textbf{Xiezhi-Meta}

We annotated 20,124 questions collected from the Graduate Entrance Examination to form the meta version of Xiezhi through both manual efforts and chatGPT.
The aim of annotation is to remove unanswerable questions and to tag each question with as many disciplines as possible.

We first used ChatGPT to tag each question with first or second-level disciplines in Chinese. 
In the process of tagging, we construct a prompt by concatenating the description of a question with its options, answers, and exam information with the description of each discipline to increase chatGPT's understanding of the question so that the question could be better tagged.
The prompts we used is listed in Appendix Prompt, and the detail of the annotation process is described in Appendix Mannual Annotation.
}

\textbf{Question Generation}

Xiezhi comprises nearly 80k multiple-choice questions generated from academic surveys, as they frequently encompass well-established domain knowledge.
\myred{
We select Chinese academic papers across all disciplines that incorporate the terms ``survey'' or ``review'' in their titles.
Subsequently, we extract several longest sentences from these surveys, which typically are the introductory sentences that contain comprehensive descriptive information pertinent to a particular field of knowledge.
}
We identify keywords using the OpenNER method~\cite{zhu2019openner} from these sentences, which are then masked to formulate the questions. 
To assemble the set of options for each question, the answers to all other questions in Xiezhi were sampled and combined with the standard answers for each respective question.

\textbf{Auto Annotation}

The objectives of auto annotation include the elimination of unanswerable questions and the assignment of relevant discipline labels to each question.
For unanswerable questions, we extracted keywords from the Xiezhi-Meta, such as ``as shown in the figure below'' or ``as listed in the table'' and so on, and exclude questions that contain any of these keywords from collected data.
\myred{
We use ChatGPT and an annotation model trained by Xiezhi-Meta to do the discipline labels tagging.
The annotation model, which is based on llama-7B, is used to tag coarse-grained discipline labels~(The Discipline Categories in this paper) to the questions.
Based on the tagged coarse-grained labels, we employ ChatGPT to assign more fine-grained labels~(First and Second-level discipline labels) to the questions, in a similar manner to the labeling of Xiezhi-Meta.
The detail about the training process of the annotation model and the performance of the auto annotation process is described in Appendix Auto Annotator.
}


\subsubsection{Xiezhi-Specialty \& Xiezhi-Interdiscipline}
\begin{figure*}[h]
    \begin{center}
    \resizebox{\textwidth}{!}{
    \includegraphics{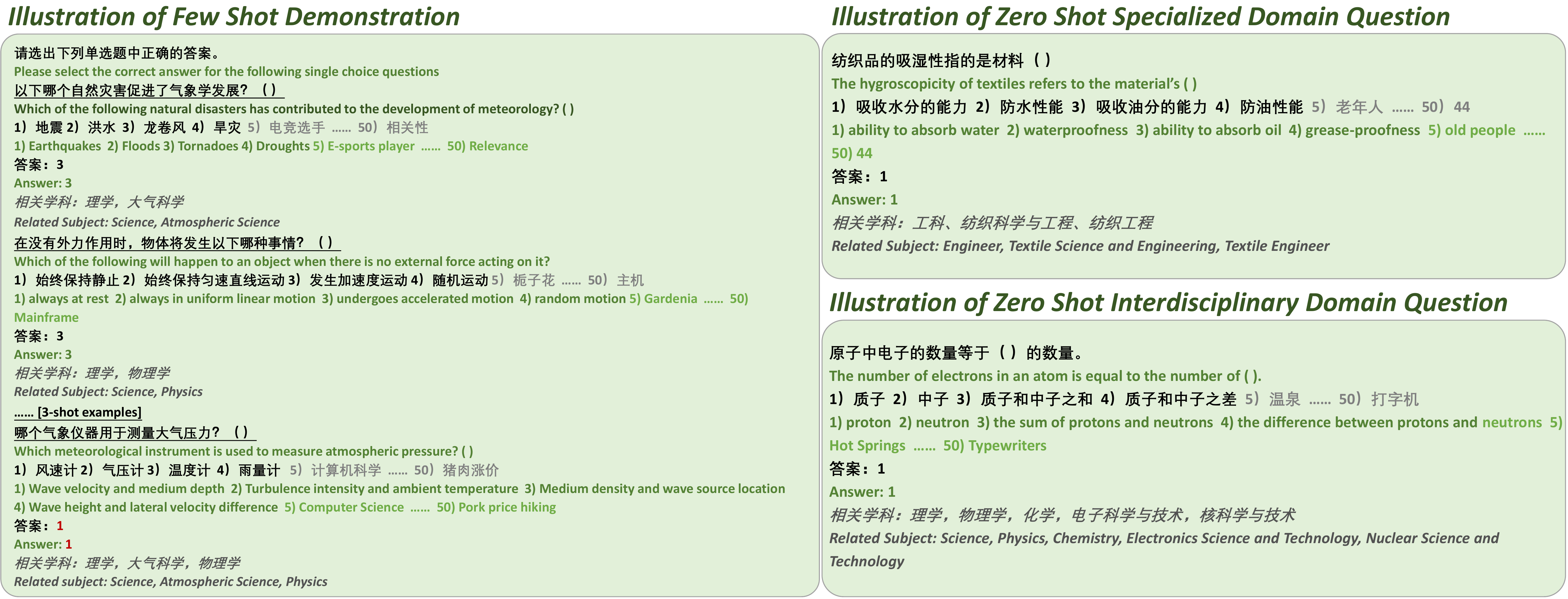}
    }
    \end{center}
    
    \caption{
    Examples of a 3-shot evaluation with Xiezhi-Interdiscipline, a question from Xiezhi-Interdiscipline and a question from Xiezhi-Specialty.
    The red text in the left figure is the autocompleted response from the model, while the preceding text is the inputted prompt.
    English translations are shown below the corresponding Chinese text for better readability.
    }
    \label{fig:questions}
\end{figure*}

To ensure the validity of the evaluation results, we further propose two additional datasets, Xiezhi-Specialty and Xiezhi-Interdiscipline \myred{in both Chinese and English version}. 
The trajectory of LLM development tends to consolidate multiple capabilities within individual LLMs, which may consequently yield unanticipated interdisciplinary problem-solving proficiencies.
The division of Xiezhi into the Specialty and Interdiscipline datasets is designed to correspond with this evolving trend.
These datasets are derived from the original Xiezhi Benchmark with the exclusion of some sensitive questions (e.g., military science) and deeply Chinese-centric questions (e.g., Literary Chinese QA, ancient Chinese poetry completion).
Based on a balanced sampling strategy, Xiezhi-Specialty is constructed by selecting questions involved in 3 disciplines or less, while Xiezhi-Interdiscipline includes questions tagged by 4 disciplines or more.
The down-right of Fig.~\ref{fig:questions} presents an instance of the Xiezhi-Specialty, while an instance of the Xiezhi-Interdiscipline is depicted in top-right of Fig.~\ref{fig:questions}.
\myred{
The process of translation and annotation is delineated in Appendix Manual Annotation.
Furthermore, Appendix Bias, Ethical Problems and Social Impact comprehensively discusses potential ethical challenges and our effort undertaken to mitigate them.
}

\section{Experiments}
\label{sec:Experiments}
\subsection{Setup}
\label{041}

\textbf{Models\myred{\&Device}:} 
We conducted experiments on 47 cutting-edge LLMs, the detailed descriptions of all tested LLMs are listed in Tab\myred{~\ref{tab:models}} in Appendix.
Our experiments cover 45 open-source LLMs based on eight different base models: \textit{bloom, llama, moss, pythia, gpt-neox, stablelm, chatGLM} and \textit{falcon}.
Considering the legal issues, we only show the results of two publicly recognized API-based LLMs, ChatGPT and GPT-4.
\myred{Our experiment was carried out on a DGX Station with 8 80G memory Tesla A100.}

\textbf{More options:}  
All tested LLMs need to choose the best-fit answer from 50 options for each question.
Each question is set up with 3 confusing options in addition to the correct answer, and another 46 options are randomly sampled from all options in all questions in Xiezhi.
\myred{We obtain options from questions that have different discipline categories and select options that do not have any identical characters~(for Chinese) or identical 4-gram characters~(for English) to the ground truth.}
It is worth noting that it is possible to use WordNet, open source synonym databases, or other word construction methods to generate more confusing options.
However, our experiments show that the performance of all LLMs declined dramatically when the number of options increased, even when using so many non-confusing options.
This achieves our goal of exacerbating the performance gap between LLMs through new experimental settings and also shows that the traditional 4-choice setting has room for improvement.

\begin{table*}[!ht]
    \centering
    \resizebox{\textwidth}{!}{
    \begin{tabular}{c|ccc|ccc|c||ccc|ccc|ccc|ccc}
        \toprule
         \textbf{Models} & \multicolumn{3}{c|}{MMLU} & \multicolumn{3}{c|}{CEval} & \multicolumn{1}{c||}{M3KE} & \multicolumn{3}{c|}{Xiezhi-Spec.-Chinese} & \multicolumn{3}{c|}{Xiezhi-Inter.-Chinese} & \multicolumn{3}{c|}{Xiezhi-Spec.-English} & \multicolumn{3}{c}{Xiezhi-Inter.-English}\\
         & 0-shot & 1-shot & 3-shot & 0-shot & 1-shot & 3-shot & 0-shot & 0-shot & 1-shot & 3-shot & 0-shot & 1-shot & 3-shot & 0-shot & 1-shot & 3-shot & 0-shot & 1-shot & 3-shot\\
        \hline
        Random-Guess                               & 0.089 & 0.089 & 0.089 & 0.089 & 0.089 & 0.089 & 0.089 & 0.089 & 0.089 & 0.089 & 0.089 & 0.089 & 0.089 & 0.089 & 0.089 & 0.089 & 0.089 & 0.089 & 0.089 \\
        \hline
        &\multicolumn{19}{c}{Generation Probability For Ranking}\\
        \hline
        Bloomz-560m                   & 0.111 & 0.109 & 0.119 & 0.124 & 0.117 & 0.103 & 0.126   & 0.123 & 0.127 & 0.124 & 0.130 & 0.138 & 0.140 & 0.113 & 0.116 & 0.123 & 0.124 & 0.117 & 0.160 \\
        Bloomz-1b1                    & 0.131 & 0.116 & 0.128 & 0.107 & 0.115 & 0.110 & 0.082   & 0.138 & 0.108 & 0.107 & 0.117 & 0.125 & 0.123 & 0.130 & 0.119 & 0.114 & 0.144 & 0.129 & 0.145\\
        Bloomz-1b7                    & 0.107 & 0.117 & 0.164 & 0.054 & 0.058 & 0.103 & 0.102   & 0.165 & 0.151 & 0.159 & 0.152 & \textbf{0.214} & 0.170 & 0.133 & 0.140 & 0.144 & 0.150 & 0.149 & 0.209\\
        Bloomz-3b                     & 0.139 & 0.084 & 0.146 & 0.168 & \textbf{0.182} & 0.194 & 0.063   & 0.186 & 0.154 & 0.168 & 0.151 & 0.180 & 0.182 & 0.201 & 0.155 & 0.156 & 0.175 & 0.164 & 0.158 \\
        Bloomz-7b1                    & 0.167 & 0.160 & 0.205 & 0.074 & 0.072 & 0.073 & 0.073   & 0.154 & 0.178 & 0.162 & 0.148 & 0.160 & 0.156 & 0.176 & 0.153 & 0.207 & 0.217 & 0.204 & 0.229 \\
        Bloomz-7b1-mt                 & 0.189 & 0.196 & 0.210 & 0.077 & 0.078 & 0.158 & 0.072   & 0.163 & 0.175 & 0.154 & 0.155 & 0.195 & 0.164 & 0.180 & 0.146 & 0.219 & 0.228 & 0.171 & 0.232 \\
        Bloomz-7b1-p3                 & 0.066 & 0.059 & 0.075 & 0.071 & 0.070 & 0.072 & 0.081   & 0.177 & 0.198 & 0.158 & 0.183 & 0.173 & 0.170 & 0.130 & 0.130 & 0.162 & 0.157 & 0.132 & 0.134 \\
        Bloomz                        & 0.051 & 0.066 & 0.053 & 0.142 & 0.166 & \underline{\textbf{0.240}} & 0.098   & 0.185 & 0.133 & 0.277 & 0.161 & 0.099 & \textbf{0.224} & 0.069 & 0.082 & 0.056 & 0.058 & 0.055 & 0.049 \\
        Bloomz-mt                     & \underline{\textbf{0.266}} & \underline{\textbf{0.264}} & \underline{\textbf{0.248}} & \underline{\textbf{0.204}} & 0.164 & 0.151 & \textbf{0.161}   & \underline{\textbf{0.253}} & \textbf{0.198} & \textbf{0.212} & \textbf{0.213} & 0.189 & 0.184 & \underline{\textbf{0.379}} & \underline{\textbf{0.396}} & \underline{\textbf{0.394}} & \underline{\textbf{0.383}} & \underline{\textbf{0.405}} & \underline{\textbf{0.398}} \\
        Bloomz-p3                     & 0.115 & 0.093 & 0.057 & 0.118 & 0.137 & 0.140 & 0.115   & 0.136 & 0.095 & 0.105 & 0.086 & 0.065 & 0.098 & 0.139 & 0.097 & 0.069 & 0.176 & 0.141 & 0.070 \\
        \hline
        llama-7b                      & 0.125 & \textbf{0.132} & 0.093 & 0.133 & 0.106 & 0.110 & \textbf{0.158}   & \textbf{0.152} & 0.141 & 0.117 & 0.142 & 0.135 & 0.128 & 0.159 & 0.165 & 0.161 & \textbf{0.194} & 0.183 & 0.176 \\
        llama-13b                     & \textbf{0.166} & 0.079 & \textbf{0.135} & 0.152 & \textbf{0.181} & \textbf{0.169} & 0.131   & 0.133 & \underline{\textbf{0.241}} & \textbf{0.243} & \textbf{0.211} & \textbf{0.202} & \underline{\textbf{0.303}} & 0.154 & 0.183 & \textbf{0.215} & 0.174 & \textbf{0.216} & \textbf{0.231} \\
        llama-30b                     & 0.076 & 0.107 & 0.073 & 0.079 & 0.119 & 0.082 & 0.079   & 0.140 & 0.206 & 0.162 & 0.186 & \textbf{0.202} & 0.183 & 0.110 & 0.195 & 0.161 & 0.088 & 0.158 & 0.219 \\
        llama-65b                     & 0.143 & 0.121 & 0.100 & \textbf{0.154} & 0.141 & 0.168 & 0.125   & 0.142 & 0.129 & 0.084 & 0.108 & 0.077 & 0.077 & \textbf{0.183} & \textbf{0.204} & 0.172 & 0.133 & 0.191 & 0.157 \\
        \hline
        baize-7b~(lora)               & 0.129 & 0.091 & 0.079 & \textbf{0.194} & 0.180 & \textbf{0.206} & \underline{\textbf{0.231}}   & \textbf{0.216} & 0.148 & 0.123 & 0.173 & 0.158 & 0.198 & \textbf{0.182} & 0.190 & 0.194 & \textbf{0.218} & 0.188 & 0.209 \\
        baize-7b-healthcare~(lora)    & 0.130 & 0.121 & 0.106 & 0.178 & 0.174 & 0.178 & 0.203   & 0.178 & 0.146 & 0.123 & \underline{\textbf{0.266}} & 0.107 & 0.118 & 0.175 & 0.164 & 0.173 & 0.197 & \textbf{0.231} & 0.198 \\
        baize-13b~(lora)              & 0.131 & 0.111 & 0.171 & 0.184 & 0.178 & 0.195 & 0.155   & 0.158 & \textbf{0.221 }& \underline{\textbf{0.256}} & 0.208 & 0.200 & \textbf{0.219} & 0.176 & 0.189 & \textbf{0.239} & 0.187 & 0.185 & \textbf{0.274} \\
        baize-30b~(lora)              & \textbf{0.193} & \textbf{0.216} & \textbf{0.207} & 0.191 & \underline{\textbf{0.196}} & 0.121 & 0.071   & 0.109 & 0.212 & 0.190 & 0.203 & \underline{\textbf{0.256}} & 0.200 & 0.167 & \textbf{0.235} & 0.168 & 0.072 & 0.180 & 0.193 \\
        \hline
        Belle-0.2M                    & 0.127 & \textbf{0.148} & \textbf{0.243} & 0.053 & 0.063 & \textbf{0.136} & \textbf{0.076}   & 0.172 & 0.126 & 0.153 & 0.171 & 0.165 & 0.147 & 0.206 & 0.146 & 0.148 & \textbf{0.217} & 0.150 & 0.173 \\          
        Belle-0.6M                    & 0.091 & 0.114 & 0.180 & \textbf{0.082} & \textbf{0.080} & 0.090 & 0.075   & \textbf{0.188} & 0.149 & \textbf{0.198} & \textbf{0.188} & \textbf{0.188} & 0.175 & 0.173 & \textbf{0.172} & \textbf{0.183} & 0.193 & \textbf{0.184} & \textbf{0.196} \\
        Belle-1M                      & \textbf{0.137} & 0.126 & 0.162 & 0.066 & 0.065 & 0.072 & 0.066   & 0.170 & 0.152 & 0.147 & 0.173 & 0.176 & \textbf{0.197} & \textbf{0.211} & 0.137 & 0.149 & 0.207 & 0.151 & 0.185 \\
        Belle-2M                      & 0.127 & \textbf{0.148} & 0.132 & 0.058 & 0.063 & \textbf{0.136} & 0.057   & 0.163 & \textbf{0.166} & 0.130 & 0.159 & 0.177 & 0.163 & 0.155 & 0.106 & 0.166 & 0.151 & 0.150 & 0.138 \\
        \hline
        chatglm-6B                    & \textbf{0.099} & \textbf{0.109} & \textbf{0.112} & \textbf{0.084} & 0.074 & \textbf{0.114} & \textbf{0.115}   & \textbf{0.082} & \textbf{0.097} & \textbf{0.147} & \textbf{0.104} & \textbf{0.111} & \textbf{0.144} & \textbf{0.106} & \textbf{0.120} & \textbf{0.124} & 0.099 & \textbf{0.079} & \textbf{0.097} \\
        doctorglm-6b                  & 0.093 & 0.076 & 0.065 & 0.037 & \textbf{0.085} & 0.051 & 0.038   & 0.062 & 0.068 & 0.044 & 0.047 & 0.056 & 0.043 & 0.069 & 0.053 & 0.043 & 0.106 & 0.059 & 0.059 \\
        \hline
        moss-base-16B                 & \textbf{0.072} & 0.050 & \textbf{0.062} & \textbf{0.115} & 0.048 & 0.052 & \textbf{0.099}   & \textbf{0.105} & 0.051 & 0.059 & \textbf{0.123} & 0.054 & 0.058 & \textbf{0.124} & \textbf{0.077} & \textbf{0.080} & \textbf{0.121} & 0.058 & 0.063 \\
        moss-sft-16B                  & 0.064 & \textbf{0.065} & 0.051 & 0.063 & \textbf{0.062} & \textbf{0.072} & 0.075   & 0.072 & \textbf{0.067} & \textbf{0.068} & 0.073 & \textbf{0.081} & \textbf{0.066} & 0.071 & 0.070 & 0.059 & 0.074 & \textbf{0.084} & \textbf{0.075} \\
        \hline
        vicuna-7b                     & 0.051 & 0.051 & 0.029 & 0.063 & 0.071 & 0.064 & 0.059   & 0.169 & \textbf{0.171} & 0.165 & 0.134 & \textbf{0.201} & \textbf{0.213} & 0.182 & \textbf{0.209} & \textbf{0.195} & \textbf{0.200} & \textbf{0.214} & 0.182 \\
        vicuna-13b                    & 0.109 & 0.104 & 0.066 & 0.060 & \textbf{0.131} & \textbf{0.131} & 0.067   & \textbf{0.171} & 0.167 & \textbf{0.166} & 0.143 & 0.147 & 0.178 & 0.121 & 0.139 & 0.128 & 0.158 & 0.174 & \textbf{0.191} \\
        alpaca-7b                     & \textbf{0.135} & \textbf{0.170} & \textbf{0.202} & \textbf{0.137} & 0.119 & 0.113 & \textbf{0.142}   & 0.129 & 0.139 & 0.123 & \textbf{0.178} & 0.104 & 0.097 & \textbf{0.189} & 0.179 & 0.128 & \textbf{0.200} & 0.185 & 0.149 \\
        \hline
        pythia-1.4b                   & \textbf{0.124} & \textbf{0.127} & \textbf{0.121} & \textbf{0.108} & \textbf{0.132} & \textbf{0.138} & 0.083   & \textbf{0.125} & \textbf{0.128} & \textbf{0.135} & 0.111 & 0.146 & 0.135 & \textbf{0.158} & \textbf{0.124} & \textbf{0.124} & \textbf{0.166} & 0.126 & 0.118 \\
        pythia-2.8b                   & 0.103 & 0.110 & 0.066 & 0.064 & 0.089 & 0.122 & 0.086   & 0.114 & 0.120 & 0.131 & 0.091 & 0.113 & 0.112 & 0.126 & 0.118 & 0.112 & 0.110 & \textbf{0.145} & 0.107 \\
        pythia-6.9b                   & 0.115 & 0.070 & 0.084 & 0.078 & 0.073 & 0.094 & 0.073   & 0.086 & 0.094 & 0.092 & 0.097 & 0.098 & 0.085 & 0.091 & 0.088 & 0.083 & 0.099 & 0.099 & 0.096 \\
        pythia-12b                    & 0.075 & 0.059 & 0.066 & 0.077 & 0.097 & 0.078 & \textbf{0.098}   & 0.102 & 0.126 & 0.132 & \textbf{0.125} & \textbf{0.147} & \textbf{0.159} & 0.079 & 0.098 & 0.110 & 0.094 & 0.120 & \textbf{0.120} \\
        \hline
        gpt-neox-20b                  & 0.081 & \textbf{0.132} & 0.086 & 0.086 & \textbf{0.096} & 0.069 & 0.094   & \textbf{0.140} & \textbf{0.103} & \textbf{0.109} & \textbf{0.120} & \textbf{0.098} & 0.085 & 0.088 & \textbf{0.101} & \textbf{0.116} & 0.099 & \textbf{0.113} & \textbf{0.156} \\
        h2ogpt-12b                    & 0.075 & 0.087 & 0.078 & 0.080 & 0.078 & \textbf{0.094} & 0.070   & 0.065 & 0.047 & 0.073 & 0.076 & 0.061 & \textbf{0.091} & 0.088 & 0.050 & 0.065 & 0.105 & 0.063 & 0.067 \\
        h2ogpt-20b                    & \textbf{0.114} & 0.098 & \textbf{0.110} & \textbf{0.094} & 0.084 & 0.061 & \textbf{0.096}   & 0.108 & 0.080 & 0.073 & 0.086 & 0.081 & 0.072 & \textbf{0.108} & 0.068 & 0.086 & \textbf{0.109} & 0.071 & 0.079 \\
        \hline
        dolly-3b                      & 0.066 & 0.060 & 0.055 & 0.079 & \textbf{0.083} & \textbf{0.077} & 0.066   & 0.100 & 0.090 & 0.083 & 0.091 & 0.093 & 0.085 & 0.079 & 0.063 & 0.077 & 0.076 & 0.074 & 0.084 \\
        dolly-7b                      & \textbf{0.095} & \textbf{0.068} & 0.052 & \textbf{0.091} & 0.079 & 0.070 & 0.108 & \textbf{0.108} & 0.089 & 0.092 & \textbf{0.111} & 0.095 & 0.100 & \textbf{0.096} & 0.059 & 0.086 & \textbf{0.123} & 0.085 & 0.090 \\
        dolly-12b                     & \textbf{0.095} & \textbf{0.068} & \textbf{0.093} & 0.085 & 0.071 & 0.073 & \textbf{0.114} & 0.098 & \textbf{0.106} & \textbf{0.103} & 0.094 & \textbf{0.114} & \textbf{0.106} & 0.086 & \textbf{0.088} & \textbf{0.098} & 0.088 & \textbf{0.102} & \textbf{0.116} \\
        \hline
        stablelm-3b                   & 0.070 & 0.085 & 0.071 & 0.086 & 0.082 & \textbf{0.099} & 0.096   & \textbf{0.101} & 0.087 & 0.091 & 0.083 & 0.092 & 0.067 & 0.069 & 0.089 & 0.081 & 0.066 & 0.085 & 0.088 \\
        stablelm-7b                   & \textbf{0.158} & \textbf{0.118} & \textbf{0.093} & \textbf{0.133} & \textbf{0.102} & 0.093 & \textbf{0.140}   & 0.085 & \textbf{0.118} & \textbf{0.122} & \textbf{0.123} & \textbf{0.130} & \textbf{0.095} & \textbf{0.123} & \textbf{0.103} & \textbf{0.100} & \textbf{0.134} & \textbf{0.121} & \textbf{0.105} \\
        \hline
        falcon-7b                     & 0.048 & 0.046 & 0.051 & 0.046 & 0.051 & 0.052 & 0.050   & 0.077 & \textbf{0.096} & \textbf{0.112} & \textbf{0.129} & \textbf{0.141} & \textbf{0.142} & 0.124 & 0.103 & 0.107 & \textbf{0.198} & \textbf{0.200} & \textbf{0.205} \\
        falcon-7b-instruct            & 0.078 & 0.095 & 0.106 & \textbf{0.114} & \textbf{0.095} & 0.079 & 0.104   & 0.075 & 0.083 & 0.087 & 0.060 & 0.133 & 0.123 & \textbf{0.160} & \textbf{0.203} & \textbf{0.156} & 0.141 & 0.167 & 0.152 \\
        falcon-40b                    & 0.038 & 0.043 & 0.077 & 0.085 & 0.090 & \textbf{0.129} & 0.087   & 0.069 & 0.056 & 0.053 & 0.065 & 0.063 & 0.058 & 0.059 & 0.077 & 0.066 & 0.085 & 0.063 & 0.076 \\
        falcon-40b-instruct           & \textbf{0.126} & \textbf{0.123} & \textbf{0.121} & 0.070 & 0.080 & 0.068 & \textbf{0.141} & \textbf{0.103} & 0.085 & 0.079 & 0.115 & 0.082 & 0.081 & 0.118 & 0.143 & 0.124 & 0.083 & 0.108 & 0.104 \\
        \hline
        &\multicolumn{19}{c}{Instruction For Ranking}\\
        \hline
        GPT-3.5                       & 0.240 & 0.298 & 0.371 & 0.286 & 0.289 & 0.360 & 0.290   & 0.218 & 0.352 & 0.414 & 0.266 & 0.418 & 0.487 & 0.217 & 0.361 & 0.428 & 0.305 & 0.452 & 0.517  \\
        GPT-4                         & 0.402 & 0.415 & 0.517 & 0.413 & 0.410 & 0.486 & 0.404   & 0.392 & 0.429 & 0.490 & 0.453 & 0.496 & 0.565 & 0.396 & 0.434 & 0.495 & 0.463 & 0.506 & 0.576  \\
        \hline
        &\multicolumn{19}{c}{Statistic}\\
        \hline
        Performance-Average           & 0.120 & 0.117 & 0.125 & 0.113 & 0.114 & 0.124 & 0.111 & 0.140 & 0.140 & 0.145 & 0.144 & 0.148 & 0.152 & 0.145 & 0.145 & 0.150 & 0.156 & 0.157 & 0.166 \\
        Performance-Variance          & 0.062 & 0.068 & 0.087 & 0.067 & 0.065 & 0.078 & 0.064 & 0.058 & 0.070 & 0.082 & 0.067 & 0.082 & 0.095 & 0.067 & 0.080 & 0.090 & 0.078 & 0.092 & 0.104 \\
        \bottomrule
    \end{tabular}
    }
    \caption{
    The overall performance of all language models are listed in this table. All tested models are divided into broad groups sharing similar character. Bolded font indicates the best performing result within a group, and underlined font indicates the best performing result for the same data set with the same settings.
    }
    \label{tab:all-llm}
\end{table*}

\textbf{Few-Shot Demonstration:}  
Additionally, we aim to test the LLMs' understanding of demonstrations. 
Therefore, we evaluate the LLMs' capabilities under 0-shot, 1-shot, and 3-shot settings.
Although previous researches use a 5-shot setting, our experiments have much bigger options number for each question, taking the maximum input length of each LLM into consideration, we only use at most 3 examples in our few-shot learning experiments.
The examples used for demonstration were obtained from Xiezhi-Train, a dataset containing 2,555 questions absent from Xiezhi-Speciality and Xiezhi-Interdiscipline, with a minimum of two labels matching the test questions, an illustration is depicted in Fig.~\ref{fig:questions}.

\textbf{Metrics:}  
In this section, we present mainly two experiment results: the overall performance of all LLMs across various benchmarks, and the ranking of the top eight 0-shot LLMs in 12 non-sensitive domain categories of the Xiezhi-Benchmark with the scores for top and average practitioners.
For the 45 open-source models assessed in our evaluation, we calculated the probability of each model choosing every option using generative probabilities and then ranked all options accordingly based on the probabilities.
Due to legal considerations, we only display the results of two publicly recognized API-based LLMs: ChatGPT and GPT-4, and we ask them to rank all given options through instructions.
To represent the results of all ranking outcomes, we employed the Mean Reciprocal Rank (MRR) as the metric \myred{in this section}, which calculates the reciprocal rank of the correct answer.
MRR closer to 1 indicates that the model is more capable of placing the correct answer at the front of the ranking, while it suggests that the LLM tends to place the correct answer at the bottom if it is closer to 0.
\myred{As a comparison, we also employ four different metrics and detailed them in Appendix Results on Other Metrics}.

\myred{
\textbf{Randomness:}
To reduce the effect of randomness on our experiment,
we set the random seed of some python libraries used in our experiment, which are \texttt{Numpy}, \texttt{Random}, and \texttt{Torch}, to 42.
It is worth noting that since we used a generative probability to rank each option, this generative probability is independent of the hyperparameters to each LLMs.
Nonetheless, in order to be consistent in our experiments even for details we did not notice, we still set the deterministic hyperparameters, as described in Appendix Detail Hyper-parameters.
Besides, Given that each question need to sample other 46 options, we constructed the set of options for each question before we started our experiment to ensure the consistency in our experiment.
Also, we used string similarity during sampling to select questions that were very unlikely to be standard answers.

\begin{table*}[!ht]
    \centering
    \resizebox{\textwidth}{!}{
    \begin{tabular}{c|cc|cccccccc}
        \toprule
        \multirow{2}{*}{Category} & \multicolumn{2}{c|}{Human} & \multicolumn{8}{c}{\multirow{2}{*}{Language Models}}
        \\
        & Top & Average & \\
        \hline
          \multirow{2}{*}{Philosophy} & \multirow{2}{*}{0.856\cmark} & \multirow{2}{*}{0.453\xmark} & ChatGPT & bloomz-mt & GPT-4 & pythia-1.4b & llama-7b-hf & BELLE-7B-0.2M & BELLE-7B-1M & vicuna-13b-delta-v1.1 \\
        &&& 0.477 & 0.453 & 0.413 & 0.321 & 0.241 & 0.228 & 0.226 & 0.223 \\
         \multirow{2}{*}{Economics} & \multirow{2}{*}{0.871\cmark} & \multirow{2}{*}{0.520\cmark} & GPT-4 & bloomz-mt & llama-65b-hf & BELLE-7B-1M & llama-7b-hf & falcon-7b & baize-lora-7B & falcon-7b-instruct \\
        &&& 0.419 & 0.310 & 0.290 & 0.255 & 0.234 & 0.233 & 0.222 & 0.214 \\
         \multirow{2}{*}{Jurisprudence} & \multirow{2}{*}{0.761\cmark} & \multirow{2}{*}{0.460\cmark} & GPT-4 & llama-65b-hf & baize-lora-7B & BELLE-7B-0.2M & ChatGPT & llama-7b-hf & BELLE-7B-1M & alpaca-lora-7b \\
        &&& 0.368 & 0.323 & 0.230 & 0.217 & 0.213 & 0.210 & 0.199 & 0.192 \\
         \multirow{2}{*}{Pedagogy} & \multirow{2}{*}{0.854\cmark} & \multirow{2}{*}{0.510\cmark} & GPT-4 & bloomz-mt & ChatGPT & BELLE-7B-0.2M & baize-lora-13B & pythia-1.4b & llama-65b-hf & BELLE-7B-1M\\
        &&& 0.472 & 0.442 & 0.280 & 0.251 & 0.244 & 0.241 & 0.237 & 0.237 \\
         \multirow{2}{*}{Literature} & \multirow{2}{*}{0.825\cmark} & \multirow{2}{*}{0.560\cmark} & GPT-4 & bloomz-mt & baize-healthcare-lora-7B & baize-lora-13B & baize-lora-7B & alpaca-lora-7b & BELLE-7B-0.2M & bloomz-3b \\
        &&& 0.417 & 0.405 & 0.284 & 0.249 & 0.213 & 0.206 & 0.194 & 0.187 \\
         \multirow{2}{*}{History} & \multirow{2}{*}{0.854\cmark} & \multirow{2}{*}{0.460\cmark} & GPT-4 & bloomz-mt & ChatGPT & BELLE-7B-0.2M & BELLE-7B-1M & baize-lora-7B & alpaca-lora-7b & baize-healthcare-lora-7B \\
        &&& 0.437 & 0.272 & 0.233 & 0.214 & 0.207 & 0.202 & 0.192 & 0.181 \\
         \multirow{2}{*}{Science} & \multirow{2}{*}{0.926\cmark} & \multirow{2}{*}{0.394\xmark} & GPT-4 & bloomz-mt & ChatGPT & BELLE-7B-1M & bloomz-3b & BELLE-7B-0.6M & BELLE-7B-0.2M & vicuna-7b-delta-v1.1\\
        &&& 0.436 & 0.408 & 0.220 & 0.210 & 0.200 & 0.197 & 0.191 & 0.188 \\
         \multirow{2}{*}{Engineering} & \multirow{2}{*}{0.928\cmark} & \multirow{2}{*}{0.380\xmark} & GPT-4 & ChatGPT & bloomz-mt & bloomz-7b1 & bloomz-7b1-mt & falcon-7b & alpaca-lora-7b & BELLE-7B-1M \\
        &&& 0.420 & 0.412 & 0.387 & 0.274 & 0.253 & 0.228 & 0.224 & 0.215 \\
         \multirow{2}{*}{Agronomy} & \multirow{2}{*}{0.902\cmark} & \multirow{2}{*}{0.333\xmark} & GPT-4 & bloomz-mt & ChatGPT & bloomz-7b1-mt & BELLE-7B-0.2M & bloomz-7b1 & bloomz-3b & pythia-1.4b
\\
        &&& 0.515 & 0.366 & 0.311 & 0.224 & 0.216 & 0.215 & 0.200 & 0.193 \\
         \multirow{2}{*}{Medicine} & \multirow{2}{*}{0.805\cmark}  & \multirow{2}{*}{0.430\xmark} & GPT-4 & baize-healthcare-lora-7B & ChatGPT & doctorglm-6b & BELLE-7B-0.2M & bloomz-7b1 & bloomz-7b1-mt & BELLE-7B-1M \\
        &&& 0.469 & 0.279 & 0.265 & 0.253 & 0.223 & 0.222 & 0.219 & 0.210 \\
         \multirow{2}{*}{Management} & \multirow{2}{*}{0.857\cmark} & \multirow{2}{*}{0.513\cmark} & GPT-4 & baize-lora-30B & pythia-2.8b & bloomz-p3 & BELLE-7B-0.2M & baize-lora-7B & baize-healthcare-lora-7B & BELLE-7B-1M
\\
         &&& 0.390 & 0.375 & 0.367 & 0.280 & 0.268 & 0.268 & 0.263 & 0.259 \\
         \multirow{2}{*}{Art Studies} & \multirow{2}{*}{0.821\cmark} & \multirow{2}{*}{0.400\xmark} & GPT-4 & baize-healthcare-lora-7B & bloomz-mt & ChatGPT & BELLE-7B-0.2M & baize-lora-13B & alpaca-lora-7b & moss-moon-003-base\\
         &&& 0.437 & 0.417 & 0.377 & 0.339 & 0.238 & 0.229 & 0.227 & 0.224 \\
        \hline
        Xiezhi & GPT-4 & \multicolumn{1}{c}{bloomz-mt} & \multicolumn{1}{c}{ChatGPT} & BELLE-7B-0.2M & BELLE-7B-1M & bloomz-7b1 & baize-lora-7B & bloomz-7b1-mt & alpaca-lora-7b & vicuna-7b-delta-v1.1\\
        Overall & 0.431 & \multicolumn{1}{c}{0.337} & \multicolumn{1}{c}{0.267} & 0.211 & 0.209 & 0.203 & 0.200 & 0.196 & 0.194 & 0.191\\
        
        \hline
        MMLU & GPT-4 & \multicolumn{1}{c}{Bloomz-mt} & \multicolumn{1}{c}{ChatGPT} & baize-30b~(lora) & Bloomz-7b1-mt & Bloomz-7b1 & llama-13b & stablelm-7b & llama-65b & Bloomz-3b\\
        Overall& 0.402 & \multicolumn{1}{c}{0.266} & \multicolumn{1}{c}{0.240} & 0.193 & 0.189 & 0.167 & 0.166 & 0.158 & 0.143 & 0.139\\
        
        \hline
        C-Eval & GPT-4 & \multicolumn{1}{c}{ChatGPT} & \multicolumn{1}{c}{Bloomz-mt} & baize-7b~(lora) & baize-30b~(lora) & baize-13b~(lora) & baize-7b-healthcare~(lora) & Bloomz-3b & llama-65b & llama-13b\\
        Overall& 0.413 & \multicolumn{1}{c}{0.286} & \multicolumn{1}{c}{0.204} & 0.194 & 0.191 & 0.184 & 0.178 & 0.168 & 0.154 & 0.152\\
        
        \hline
        M3KE & GPT-4 & \multicolumn{1}{c}{ChatGPT} & \multicolumn{1}{c}{baize-7b~(lora)} & baize-7b-healthcare~(lora) & Bloomz-mt & llama-7b & baize-13b~(lora) & alpaca-7b & falcon-40b-instruct & stablelm-7b\\
        Overall& 0.404 & \multicolumn{1}{c}{0.290} & \multicolumn{1}{c}{0.231} & 0.203 & 0.161 & 0.158 & 0.155 & 0.142 & 0.141 & 0.140\\
        
        \bottomrule
    \end{tabular}
    }
    \caption{Ranking of all LLMs in each category in 0-shot setting. 
    \cmark\ denotes human performance exceeds the state-of-the-art LLMs, whereas \xmark\ signifies LLMs have surpassed human performance.}
    \label{tab:llm-rank}
\end{table*}


\textbf{Human Performance:}
Since we mainly collected questions from some of the most important examinations in China, whose average scores will be released annually. 
Furthermore, for various academic entrance examinations, each institution will publish the average score of their recruit students.
We annotate each question using the average score of the available corresponding examination and calculated the mean of all the questions within the benchmark where examination scores can be obtained.
Additionally, we used the average scores publicized by several of China's top institution as a representation of a higher level of human performance.
While this scoring method has its limitations, which we thoroughly analyze in Appendix Bias, Ethical Problems and Social Impact, it still provides usable human baselines for Xiezhi.
}

\subsection{Results of LLMs}
The overall performance towards Xiezhi and baselines of all LLMs are listed in Tab.~\ref{tab:all-llm}.
The ranking of all LLMs in each domain category is listed in Tab.~\ref{tab:llm-rank}.
And here we give the most intriguing observation in the experiments.

Note: 
(1) The results of GPT-4 and ChatGPT are acquired through instructions, their real capabilities of them may be higher than the score listed in the tables.
(2) Tab.~\ref{tab:llm-rank} displays the optimal outcomes, which are combined performance of Xiezhi-Specialty and Xiezhi-Interdiscipline, in both Chinese and English Xiezhi.
(3) At the moment of writing this paper, M3KE has solely released its training dataset. So we employed this dataset for conducting the experiments, which allowed us to execute only 0-shot experimental setups.

\textbf{Observation 1: Best Performance = Pretraining + Finetuning}
Examining the overall results presented in Tab.~\ref{tab:llm-rank}, it is observed that all top-10 open-source LLMs are built upon either the llama or bloom frameworks. 
This suggests that obtaining the most exceptional performance is more likely through these two base models, due to their substantial potential and superior performance in domain text comprehension. 
Moreover, it is noted that all open-source models within the top-10 overall performance in Tab.~\ref{tab:llm-rank} are finetuned models, which implies that only finetuned LLMs can attain the highest performance. 
As a result, both effective pretraining and fine-tuning processes are crucial components in attaining optimal performance in domain text comprehension.


\textbf{Observation 2: Most LLMs are incapable of performing stably few-shot learning from demonstrations}
As shown in the ``Performance-Average'' in Tab.~\ref{tab:all-llm}, the average performance of LLMs reveals that more quantity of examples results in better model performance. 
However, it is not an absolute guarantee that each LLM will exhibit enhanced performance in response to an increased number of demonstrations.
On the contrary, several LLMs exhibit a decline in performance as the quantity of learning examples expands. 
In contrast, GPT-4 and ChatGPT demonstrate a more stable improvement in their performance through few-shot learning. 
This can be attributed to the extensive domain knowledge possessed by GPT-4 and ChatGPT, enabling them to effectively comprehend the features embedded within the learning samples.

\textbf{Observation 3: More LLMs' parameters don't guarantee better performance}
Numerous studies have posited that an increase in the number of model parameters corresponds to an enhancement in model's performance. 
This notion holds true when comparing LLMs that exhibit an order of magnitude difference in their parameters. 
For instance, Bloomz-mt with 146 billion parameters significantly outperforms Bloomz-560m with 560 million parameters. 
However, this argument does not consistently hold. 
For instance, Bloomz-7b1 surpasses Bloomz-p3 in the majority of domain tasks, and Pythia-1.4b outperforms other Pythia models with larger parameter counts across most benchmarks. 
A possible explanation for this phenomenon could be that LLMs with different parameter quantities are optimally suited to different amounts of pre-training and fine-tuning data~\cite{hoffmann2022empirical}.

\textbf{Observation 4: Small LMs enhance domain capabilities at the expense of generic capabilities}
In our experiments, we examined two medical LLMs: DoctorGLM and Baize-Healthcare. 
DoctorGLM originated from ChatGLM-6B, and Baize-Healthcare was derived from Llama-7B, with both models fine-tuned using medical domain text. 
Although both models have lower MRR compared to other LLMs fine-tuned based on the same base models, they each demonstrate high performance in medical domain. 
This suggests the augmentation of LLMs with fewer parameters in domain text comprehension, whether finetuned through exclusively domain-specific data or combining domain-specific and generic data, will inevitably lead to a trade-off in the understanding of generic text. 
This observation aligns with the findings from previous research~\cite{fu2023specializing, zhao2023survey}.

\subsection{Results of Benchmarks}
Based on the observations from Tab.~\ref{tab:llm-rank}, although the objective is to comprehensively evaluate the domain capabilities of LLMs, the various benchmarks still exhibit differing results, which indicates the different emphases of each benchmark. 
GPT-4, ChatGPT, and Bloomz-mt consistently rank within the top 10 across all four benchmarks, Baize-7b, and Bloomz-7b1 demonstrate remarkable abilities as they rank within the top 10 across three of the benchmarks.
Furthermore, Xiezhi exhibits the highest variance among all LLMs in the "Performance-Variance" of Tab.~\ref{tab:all-llm}, \myred{while the score of GPT-4 doesn't always rank first like it was in other benchmark works.}
This indicates that the Xiezhi Benchmark excels at discerning the competence disparities among diverse LLMs and possesses the potential to appraise more potent LLMs.

\section{Conclusion}
We introduced Xiezhi, a new benchmark that measures how well LLMs acquire and apply domain knowledge.
By covering 516 subjects ranging from 13 categories with 249,587 questions, Xiezhi proposes a taxonomy of all human knowledge and assesses language understanding of the cutting-edge 47 LLMs in greatest breadth and depth among all previous benchmarks.
\myred{
Our research has revealed that the SOTA LLMs have outperformed practitioner experts in several domains when evaluated by multiple-choice question answering tasks. 
Furthermore, there is still a big gap in generic domain knowledge comprehension between larger and smaller models.
Our experimental findings and the Xiezhi Benchmark we developed provide researchers with a more comprehensive understanding of their capabilities across diverse domains.
}

\section{Acknowledgement}
This work was supported by 
Science and Technology Commission of Shanghai Municipality Grant (No. 22511105902).
Shanghai Municipal Science and Technology Major Project (No.2021SHZDZX0103).
National Natural Science Foundation of China (No.62102095).
National Natural Science Foundation of China (No. 62306112).
National Natural Science Foundation of China (No. U23A20496).


\bibliography{aaai24}

\newpage
\appendix
\section{Appendix}
\subsection{Discussion}
\label{sec:discussion}
\textbf{Large Language Models Need More Benchmarks}

In summary, the capabilities of LLMs are manifested in three distinct aspects~\cite{ouyang2022training}. 
And all three of these categories require benchmarks for automated evaluation.
Although many benchmarks are constructed after the release of ChatGPT or GPT-4, LLMs still faced the problem of insufficient evaluation dimensions and insufficient evaluation detail because LLMs are more expressive than ever.
Thus, we call upon the academic and industrial sectors to summarize human knowledge and values, providing LLM development with more effective, comprehensive, and advanced benchmarks.

The first capability of LLMs is the understanding of knowledge, which encompasses memorization, reasoning, and abstraction~\cite{zhou2023comprehensive}. 
Currently, most works focus on enhancing the knowledge and understanding of LLMs through pre-training~\cite{fu2022does}. 
The proposal of Xiezhi is aiming at establishing a taxonomy for human knowledge and building evaluation criteria for this field. 
Although Xiezhi is already the most dimensional domain evaluation benchmark with largest volume of data, we currently offer only Chinese and English language version and lacks comprehensive coverage of knowledge from different cultures and industries. 
In the future, one of the critical improvements for Xiezhi lies in collecting more thorough and in-depth knowledge from various countries, nations, fields, and open source benchmarks in more languages.

Except for knowledge evaluation, there are two other capabilities of LLMs that are in great need of benchmarks.
One capacity is to understand and execute instructions, rendering LLM into a valuable artificial tool~\cite{aribandi2021ext5, hoffmann2022empirical}.
Instruction fine-tuning is greatly involved in many works to enhance LLM's instruction-following ability.
However, the evaluation of LLM functionality largely relies on manual verification at present.
Another is to align with human values, which is essential for LLMs to evolve into artificial general intelligence (AGI)~\cite{bai2022constitutional,perez2022discovering}. 
Numerous technical approaches for alignment have been proposed by companies like OpenAI and Claude, but many works have not aligned their models with human values due to the lack of direct improvement in downstream applications.

\textbf{Large Language Models Need Better Evaluation Methods}

Current language models predominantly adopt generative approaches~\cite{zhao2023survey}, and naturally, assessing these models presents inherent challenges~\cite{wang2023chatgpt}. 
Most existing evaluation methods utilize multiple-choice questions to measure a generative model's understanding of knowledge and employ extraction techniques to obtain the model's answers~\cite{huang2023ceval, liu2023m3ke, hendrycks2021measuring}.

We argue that this evaluation approach is a sub-optimal approach. 
Since this approach requires models to possess the capability to answer multiple-choice questions, a skill seldom employed in real-world applications.
For small LLMs or LLMs that have not been fine-tuned with multiple-choice data, such evaluation approaches fail to provide effective performance indicators.

In this paper, we propose evaluating models by using generative probability. 
While generative probability increases computational costs in comparison to directly answering questions, it yields a more accurate and effective assessment for LLMs unable to answer multiple-choice questions. 
Our study serves as an exploration of improved evaluation methodologies. 
In the future, we will consider incorporating a wider variety and diversity of evaluation approaches.

\myred{

\subsection{Detail Hyper-parameters}
\label{a1}
Our experiments involved two types of hyperparameters.
The first type pertains to the seeds of random numbers used in various Python libraries, while the second type refers to the hyperparameters used when invoking the AutoCausalLM class from the transformers library for generation.
The first type of hyperparameters will impact our experiments to some extent, and hence we ensured that the random number seeds were all set to 42, which are presented in Table ~\ref{tab:hyperparameter}, 
Considering that we use the generation probabilities of each options without actually generate new content, so the second type of hyperparameters would not impact our experimental results since they all effect the output from a AutoCausalLM Class. 
Nonetheless, to ensure consistency on details we might not have noticed, we configured our settings as demonstrated in Table~\ref{tab:hyperparameter}.
The code and data utilized in our study can be accessed in the CodeAndDataAppendix. Reproduction of the experiments can be achieved by simply executing the \texttt{./Tester/test.sh} file contained in our code repository.

\begin{table*}[t]
    \centering
    \resizebox{\textwidth}{!}{
    \myred{
    \begin{tabular}{|c|c|c|c|c|}
    \toprule
    \multicolumn{5}{|c|}{Random Seed} \\
    \hline
    torch.manual\_seed & torch.cuda.manual\_seed\_all & numpy.random.seed & random.seed & torch.backends.cudnn.deterministirc \\
    42 & 42 & 42 & 42 & True \\
    \hline
    \hline
    \multicolumn{5}{|c|}{AutoCausalLM} \\
    \hline
    temperature & top\_p & top\_k & num\_beams & max\_new\_token \\
    0.95 & 0.95 & 5 & 2 & 1  \\
    \bottomrule
    \end{tabular}
    }}
    \caption{\myred{All the parameter setting in our experiments.}}
    \label{tab:hyperparameter}
\end{table*}

\subsection{Prompts}
\label{a2}
We employed customized prompts in two scenarios: 
During the ``Discipline Annotation'' phase in the annotation of Xiezhi and during the ``LLMs Output'' phase as part of our experiment setting. 

During the Discipline Annotation, given the existence of over 500 discipline labels, direct input of all labels will incur huge economical expense.
So we used the Discipline Taxonomy for hierarchical annotation.
Initially, all 13 discipline categories were used to query models and determine which categories the question should belong to.
Subsequently, we selected the first-level disciplines within the chosen categories for further query, a process repeated with second-level disciplines.
This annotation strategy was used to assign discipline labels to Xiezhi-Meta courtesy of ChatGPT and Xiezhi-All via both the annotation model and ChatGPT.
The prompts used for these procedures are illustrated in the ``Chinese-Version'' section of Table~\ref{tab:prompts}, with the English version of the prompts also provided in the ``English-Version'' section of the same table for better illustration.

Regarding the model's prompt-based option output, we designed four distinctive prompts to cater to Chinese and English languages, and possible demonstrations, as depicted in the ``LLMs Output'' of Table ~\ref{tab:prompts}.

\begin{table*}[t]
    \centering
    \resizebox{\textwidth}{!}{
    \myred{
    \begin{tabular}{|l|l|}
    \toprule
         Description & Prompts \\
         
         \hline
         \multicolumn{2}{|l|}{\cellcolor{mygray} \textit{Discipline Annotation}} \\
         \hline
         \begin{tabular}[l|]{@{}l@{}}
         Chinese-Version~(Used \\in annotation process.)
         \end{tabular} 
         
         & 
         
         \begin{tabular}[l|]{@{}l@{}}
         下面我会给你几个学科标签，然后会给你一道选择题的问题、选项和答案，请你从我给你的学科名称中为这\\
         道选择题选择合适的学科标签，如果你认为没有合适的标签，你可以回答``N/A''。 \\
         学科标签： \\
         \quad xxx: yyyyyyyy \\
         \quad xxx: yyyyyyyy \\
         \quad ...... \\
         选择题问题： \\
         \quad xxxxxx \\
         选择题选项： \\
         \quad 1.xxx \\
         \quad 2.yyy \\
         选择题答案： \\
         \quad xxx \\
         请告诉我你认为这道题涉及了哪些学科标签，请严格使用给定的学科标签进行回复，不同的学科标签之间请用\\中文顿号``、''进行分割： 
         \end{tabular} 
         \\

         \hline
         \begin{tabular}[l|]{@{}l@{}}
         English-Version~(Only \\for illustration.)
         \end{tabular} & 

         \begin{tabular}[l|]{@{}l@{}}
         Below, I will provide you with some subject labels and present you with a multiple-choice question, comprising of the \\ 
         problem, options, and answer. Please allocate an appropriate subject label from the given set to this multiple-choice \\
         question. If you believe that there is no suitable label, you may reply with "N/A".
         \\
         Subject Labels: \\
         \quad xxx: yyyyyyy\\
         \quad xxx: yyyyyyy\\
         \quad ...... \\
         Question Description： \\
         \quad xxxxxx \\
         Options: \\
         \quad 1.xxx \\
         \quad 2.yyy \\
         Answer: \\
         \quad xxx \\
         Please inform me of the subject labels you believe this question encompasses. Make sure to strictly use the given subject\\ labels for your response, and separate different subject labels with the Chinese punctuation mark ``、'':
         \end{tabular}
         \\
         
         \hline
         \multicolumn{2}{|l|}{\cellcolor{mygray} \textit{LLMs Output}} \\
         \hline
         
         English-0-shot &
         \begin{tabular}[l|]{@{}l@{}}
         \#\#\# question description:
         \quad```\{question\}''' \\
         \#\#\#all options:
         \quad```\{options\}'''\\
         \#\#\# answer:
         \quad```\{answer\}'''\\
         \{eos\}
         \end{tabular} \\
         
         \hline
         
         English-few-shot &
         \begin{tabular}[l|]{@{}l@{}}
         ```\{demonstrations\}'''\\
         \#\#\# question description:
         \quad```\{question\}'''\\
         \#\#\# all options:
         \quad```\{options\}'''\\
         \#\#\# answer:
         \quad```\{answer\}'''\\
         \{eos\}
         \end{tabular} \\
         
         \hline
         
         Chinese-0-shot&
         \begin{tabular}[l|]{@{}l@{}}
         \#\#\# 问题描述:
         \quad```\{question\}'''\\
         \#\#\# 所有选项:
         \quad```\{options\}'''\\
         \#\#\# 答案:
         \quad```\{answer\}'''\\
         \{eos\} 
         \end{tabular} \\
         
         \hline
         
         Chinese-few-shot &
         \begin{tabular}[l|]{@{}l@{}}
         ```\{demonstrations\}'''\\
         \#\#\# 问题描述:
         \quad```\{question\}'''\\
         \#\#\# 所有选项:
         \quad```\{options\}'''\\
         \#\#\# 答案:
         \quad```\{answer\}'''\\
         \{eos\}
         \end{tabular}\\
         \bottomrule
    \end{tabular}
    }
    }
    \caption{\myred{All the prompt we used in both annotation and experiments.}}
    \label{tab:prompts}
\end{table*}


\subsection{Manual Annotation}
\label{a3}
\begin{figure*}[t]
    \centering
    \resizebox{\textwidth}{!}{
    \includegraphics{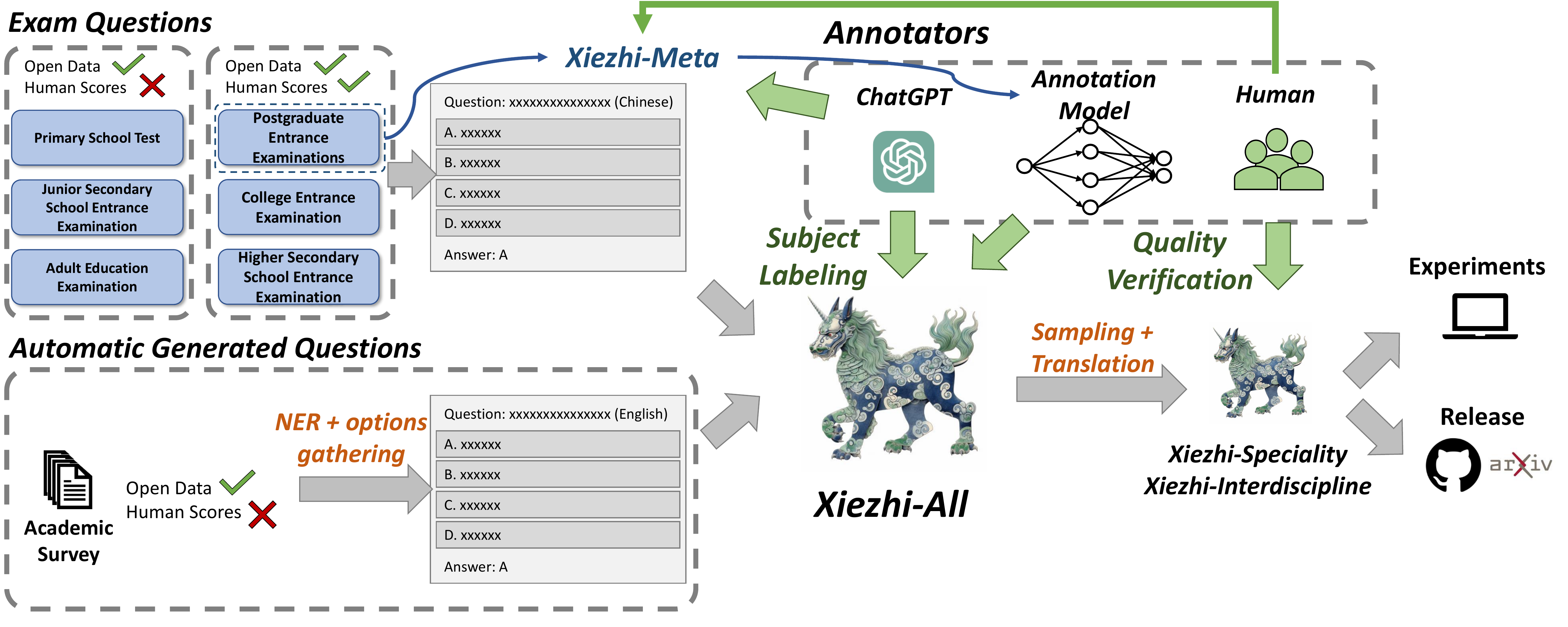}
    }
    \caption{\myred{An illustration of the construction process of Xiezhi.}}
    \label{fig:framework}
\end{figure*}

\textbf{Annotators}

We hired and paid graduate students from various majors to annotate and clean the questions we collected. 
We provided training for all potential annotators and tested their understanding of the training content to ensure they fully met our requirements.
The training mainly about the requirements in annotation, we will talk about these requirements in the next subsection.
Our annotators comprised of ten Chinese graduate students specializing in diverse disciplines: Medicine, Literature, Economics, Science, Jurisprudence, History, Management, and Engineering.
The annotators possess expertise in their respective fields of study and have written numerous academic papers in English, thereby exhibiting a high level of proficiency in the English.
We keep communicate with them during the annotation process, and all of them are paid above the local minimum wage.
To ensure the quality of the annotations, each sample is annotated by at least three annotators.

\textbf{Annotation Process in Benchmark Construction}

As shown in Fig.~\ref{fig:framework}, we performed manual annotation at two points: the construction of Xiezhi-Meta, and the Verification of Xiezhi-Specialty and Xiezhi-Interdiscipline for different purposes.

For the construction of Xiezhi-Meta, the aim is to construct a high-quality Chinese domain knowledge dataset, covering as many discipline labels as possible.
The goal of manual annotation primarily lies in ensuring the correctness of the questions themselves.
Therefore, we ask annotators to filter the dataset of Xiezhi-Meta under the following requirements: 
\begin{enumerate}

    \item Choose questions where answers can be determined solely based on textual content. 
    \item Select questions with correct answers. 
    \item Choose questions with rationally set options. 
    \item The subject annotated by ChatGPT for the question is correct, or it can be modified to be correct (this requires annotators to make modifications). 
\end{enumerate}

In terms of the construction of Xiezhi-Specialty and Xiezhi-Interdiscipline, the goal is to construct a dataset that meets the requirements of domain knowledge evaluation for LLMs and conforms to human values.
Consequently, for these two datasets, we propose the following requirements for annotators:
\begin{enumerate}
    \item If the question needs non-textual information to be solved, it should be removed.
    \item If issue that cannot be modified was introduced during the crawling of questions, it should be removed.
    \item Questions with incorrect answers should be removed.
    \item Questions with unreasonably set options should be removed.
    \item Questions that contain gender-biased content should be removed.
    \item Questions involving sensitive content such as military matters and politics should be removed.
    \item Questions that contain China's ancient texts and contemporary political content should be removed.
    \item If the subject annotation is incorrect, it should be removed.
    \item Questions, and options that highly replicate the content of other questions should be removed.
    \item Questions with discriminatory content should be removed.
    \item If deletion of a question leads to imbalance, new questions should be re-selected from Xiezhi-All to be included.
    \item The reference of male and female appellations in the dataset should be balanced.
\end{enumerate}




\textbf{Examples in Annotation Process}

In this section, we provide a detailed list of real examples encountered during our manual annotation process.
Firstly, we removed all the disciplines included in Tab.~\ref {tab:delete_subjects} from Xiezhi-Specialty and Xiezhi-Interdiscipline.
Some of these disciplines require a super deep understanding of the Chinese cultural, which does not align with current model development needs and is too difficult even for Chinese LLMs.
Others involve too much military-related content, which is not suitable for open source dissemination.
In addition, regarding all the requirements we outlined in the prior section, specific examples are shown in Table ~\ref {tab:example_in_annotation}.


\begin{table*}[t]
\centering
\resizebox{\textwidth}{!}{
\myred{
\begin{tabular}{|l|}
\hline
\cellcolor{mygray}\textbf{\textit{Deeply Chinese Related Disciplines}} \\
\hline
Chinese Classical Literature, History of the Chinese Communist Party, Ancient Chinese Literature\\
Marxist Theory and Ideological and Political Education \\
\hline
 \cellcolor{mygray}\textit{\textbf{Military Related Disciplines}} \\
 \hline
 Military Science, Military Political Work, Military Political Work Studies, \\ 
 Military Logistics, Military logistics and military, Military Strategy, Military tactics, Contract Tactics, \\
 War Mobilisation equipment science, Military Logistics, Rear professional logistics, Strategic Studies, \\
 Military Equipment Studies, Military Thought and Military History, Military Thought, Tactics, \\
 Military History, Military Equipment Studies, Military Training, Military Systems, Joint Warfare, \\
 Military Organization, Military Management, Military Command, Operational Command,\\
 Military Operations Research, Military Communications, Military Intelligence, Cryptology, \\
 Military Education and Training, Campaign Studies, Military Service Campaign Studies. \\
\hline
\end{tabular}
}}
\caption{
\myred{
All the subjects we delete from Xiezhi-Speciality and Xiezhi-Disciplinary.
}}
\label{tab:delete_subjects}
\end{table*}

\begin{table*}[t]
\centering
\resizebox{\textwidth}{!}{
\myred{
\begin{tabular}{|l|l|}
\hline
\multicolumn{2}{|l|}{\cellcolor{mygray} \textbf{\textit{Xiezhi-Meta}}}\\
\hline

\begin{tabular}[c]{@{}l@{}}
1. Do not select questions that cannot be answered based solely on \\
the content of the text.
\end{tabular}
& 
\begin{tabular}[c]{@{}l@{}}
Exclude \\
\quad Question: \\
\quad \quad ``Based on the given figure, ....''
\end{tabular}
\\

\hline

\begin{tabular}[c]{@{}l@{}}
2. Do not select questions with incorrect answers.
\end{tabular}
& 
\begin{tabular}[c]{@{}l@{}}
Exclude \\
\quad Question:  \\ 
\quad \quad ``The theory of time is first proposed by \_\_\_\_\_\_'' \\
\quad Answer: \\ 
\quad \quad ``Stephen Hawking'', 
\quad Real Answer: 
\quad \quad ``Albert Einstein''
\end{tabular}
\\

\hline

\begin{tabular}[c]{@{}l@{}}
3. Do not select questions with improperly formed alternatives.
\end{tabular}
& 
\begin{tabular}[c]{@{}l@{}}
Exclude \\
\quad Question:  \\
\quad \quad ``Who wrote `Pride and Prejudice'?'' \\ 
\quad Options:  \\
\quad \quad ``Joanne Rowling, J.K. Rowling, Stephen King, ...''
\end{tabular}
\\ 

\hline

\begin{tabular}[c]{@{}l@{}}
4. Do not select questions where the subject markers are mislabelled \\
and cannot be corrected.
\end{tabular}
& 
\begin{tabular}[c]{@{}l@{}}
Exclude \\
\quad Question:  \\ ``If you go back to the Spring and Autumn Period and the Warring States Period, \\ what you can do is \_\_\_\_\_\_ '' \\
\quad Labels: \\
\quad \quad [Ancient Literature, ...]
\end{tabular}
\\



\hline
\multicolumn{2}{|l|}{\cellcolor{mygray} \textbf{\textit{Xiezhi-Speciality \& Xiezhi-Interdiscipline}}} \\
\hline

\begin{tabular}[c]{@{}l@{}}
1. If the question needs non-textual information to be solved, \\
it should be removed.
\end{tabular}&
\begin{tabular}[c]{@{}l@{}}
Delete \\
\quad Question: \\
\quad \quad ``Based on the given figure, ....''
\end{tabular}
 \\

\hline

\begin{tabular}[c]{@{}l@{}}
2. If issue that cannot be modified was introduced during the \\
 crawling of questions, it should be removed.
\end{tabular}&
\begin{tabular}[c]{@{}l@{}}
Delete \\
\quad Question:\\
\quad \quad ``<banana> swirling css \{ <chairs margin=1em auto> \} upwardly sky blues.''
\end{tabular}
\\

\hline

\begin{tabular}[c]{@{}l@{}}
3. Questions with incorrect answers should be removed.
\end{tabular}& 
\begin{tabular}[c]{@{}l@{}}
Delete \\
\quad Question:  \\ 
\quad \quad ``The theory of time is first proposed by \_\_\_\_\_\_'' \\
\quad Answer: \\ 
\quad \quad ``Stephen Hawking'',\\ 
\quad Real Answer: \\
\quad \quad ``Albert Einstein''
\end{tabular}
 \\ 

\hline

\begin{tabular}[c]{@{}l@{}}
4. Questions with unreasonably set options should be removed.
\end{tabular}& 
\begin{tabular}[c]{@{}l@{}}
Delete \\
\quad Answer: \\ 
\quad \quad ``Samurai'',\\
\quad Options:  \\ 
\quad \quad ``Bushi, Anchor, Archer, ......''
\end{tabular}
 \\

\hline

\begin{tabular}[c]{@{}l@{}}
4. If the options are set improperly in the question,  \\
it should be deleted.
\end{tabular}& 
\begin{tabular}[c]{@{}l@{}}
\quad Question:  \\ 
\quad \quad ``Who wrote `Pride and Prejudice'?'' \\
\quad Options:  \\ 
\quad \quad ``Joanne Rowling, J.K. Rowling, Stephen King, ...''
\end{tabular}
\\ 

\hline

\begin{tabular}[c]{@{}l@{}}
5. Questions that contain gender-biased content should be removed.
\end{tabular}& 
\begin{tabular}[c]{@{}l@{}}
Delete \\
\quad Question:  \\ 
\quad \quad ``Since men usually cook better than women ...''
\end{tabular}
\\

\hline

\begin{tabular}[c]{@{}l@{}}
6. Questions involving sensitive content such as military matters \\
and politics should be removed.
\end{tabular}& 
\begin{tabular}[c]{@{}l@{}}
Delete \\
\quad Question: \\
\quad \quad ``The implications of the recent nuclear agreement between ...''
\end{tabular}
\\

\hline

\begin{tabular}[c]{@{}l@{}}
7. Questions that contain China's ancient texts and contemporary \\
political content should be removed.
\end{tabular}& 
\begin{tabular}[c]{@{}l@{}}
Delete \\
\quad Question:  \\ 
\quad \quad ``The next sentences of The Dao that can be spoken is not the eternal Dao is \_\_\_\_\_\_'' 
\end{tabular}
\\

\hline

\begin{tabular}[c]{@{}l@{}}
8. If the subject annotation is incorrect, it should be removed.
\end{tabular}& 
\begin{tabular}[c]{@{}l@{}}
Delete \\
\quad Question:  \\ 
\quad \quad ``If you go back to the Spring and Autumn Period and the Warring States Period, \\what you can do is \_\_\_\_\_\_'' \\
\quad Labels: \\
\quad \quad [Ancient Literature, ...]
\end{tabular}
\\ 

\hline

\begin{tabular}[c]{@{}l@{}}
9. Questions, and options that highly replicate the content \\
of other questions should be removed.
\end{tabular}& 
\begin{tabular}[c]{@{}l@{}}
/
\end{tabular}
\\

\hline

\begin{tabular}[c]{@{}l@{}}
10. Questions with discriminatory content should be removed.
\end{tabular}& 
\begin{tabular}[c]{@{}l@{}}
Delete \\
\quad Question:  \\ 
\quad \quad ``People of a certain race smarter than others, ...''
\end{tabular}
\\

\hline

\begin{tabular}[c]{@{}l@{}}
11. If deletion of a question leads to imbalance, new questions \\
should be re-selected from Xiezhi-All to be included.
\end{tabular}& 
\begin{tabular}[c]{@{}l@{}}
/
\end{tabular}
\\

\hline

\begin{tabular}[c]{@{}l@{}}
12. The reference of male and female appellations in the dataset \\
should be balanced.
\end{tabular}& 
\begin{tabular}[c]{@{}l@{}}
Replace the male appellations in the sentence with female ones, while paying attention \\ to modification of personal names.
\end{tabular}
\\

\hline
\end{tabular}
}
}
\caption{
\myred{
Real examples in our annotation process.
}}
\label{tab:example_in_annotation}
\end{table*}

\subsection{Auto Annotator}
\label{a4}

\begin{figure*}[h]
    \centering
    \resizebox{\textwidth}{!}{
    \includegraphics{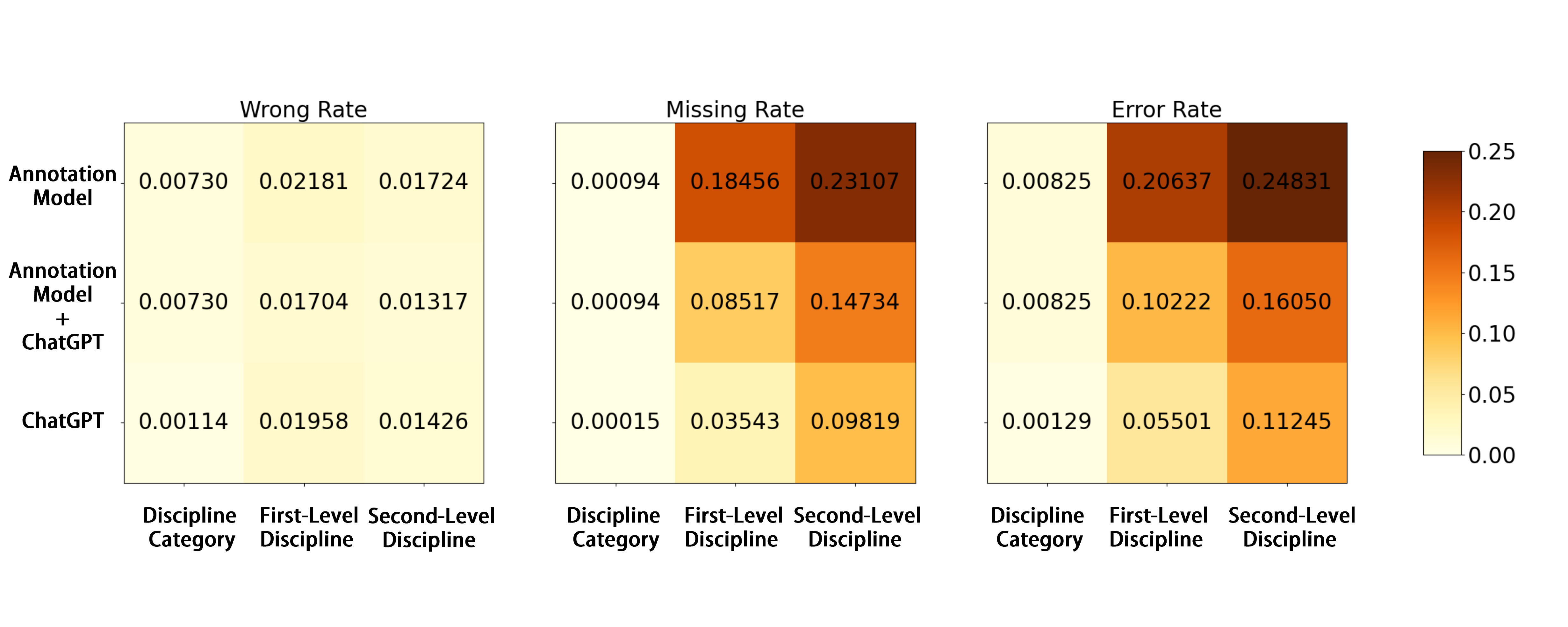}
    }
    \caption{
    \myred{
    The precision, recall, and error rates of various annotation policies.
    }}
    \label{fig:annotation}
\end{figure*}

As shown in Figure ~\ref{fig:framework}, during the construction of the benchmark, we selected 20,124 questions from the Chinese postgraduate examination to form Xiezhi-Meta. After manual verification, these translated questions were used to train an Annotation Model, which aided in annotating all data in Xiezhi-All. 
We refer to the trained model as the ``Annotation Model'', and elaborate on this model and the training details within this subsection.

\textbf{Model and Training Process:} We directly utilized the most up-to-date Llama-7B-Chinese model~\footnote{https://huggingface.co/LinkSoul/Chinese-Llama-2-7b}, which is a model enhanced through secondary pre-training using Chinese corpora based on the basic Llama-7B. Therefore, it has robust capabilities in processing Chinese Language as shown in Tab.~\ref{tab:all-llm} of the overall experiments.
We primarily fine-tuned the instruction on Llama-7B-Chinese.
The code used for this process is the EasyLLM~\footnote{https://github.com/MikeGu721/EasyLLM} training framework available on GitHub.

\textbf{Training Data: }
We constructed the data using the Chinese Version of the prompts in Table ~\ref{tab:prompts}. Although the input questions, discipline labels and descriptions of discipline labels are mainly Chinese.
We required the model to output Chinese disciplines labels relevant to the given question.

\textbf{Annotation Performance:}
Our experimental study involved the analysis of 20,124 questions derived from Xiezhi-Meta. 
Three distinct strategies, namely Annotation Model, Annotation Model + ChatGPT, and ChatGPT, were employed to annotate these questions. The annotated datasets were subsequently compared to the manual validation results. 
In the Annotation Model + ChatGPT strategy, the Annotation Model was employed to annotate the discipline categories, and ChatGPT was applied to annotate the first and second-level disciplines by leveraging the annotated discipline categories. 
The evaluation metrics used were the Wrong Rate, Missing Rate, and Error Rate:
\begin{itemize}
    \item \textbf{Wrong Rate} indicates the number of incorrect annotated discipline labels, calculated as ``[\texttt{SET}(manual labeling results) - \texttt{SET}(annotation results)]/\#questions''.
    \item \textbf{Missing Rate} is used to indicate the number of discipline labels omitted by the annotation strategy; it was calculated as ``[\texttt{SET}(annotation results) - \texttt{SET}(manual labeling results)]/\#questions''.
    \item \textbf{Error Rate} is the summation of the Wrong Rate and Missing Rate metrics, which denotes the probability of manual involvement in the annotation process.
\end{itemize}
It is important to note that multiple labels may be missed or incorrectly labeled for each question, so all the rates used here are not 0~1 metrics.

The results presented in Table~\ref{fig:annotation} reveal that the Annotation Model has good performance in coarse-grained discipline classification but is not as effective in fine-grained discipline classification.
It is observed that all three strategies aimed to ensure high precision in their outputs, resulting in low Wrong Rates.
However, the Annotation Model missed many discipline labels, particularly in the first and second-level subjects.
The Missing Rate of the Annotation Model + ChatGPT policy is much higher is because the missiong of the discipline categories annotated by Annotation Model

\textbf{Annotation of Xiezhi-All:} 
 To ensure the quality of Xiezhi-All, we applied a combined annotation approach using Annotation Model and ChatGPT.
Specifically, we used the Annotation Model to annotate the subject categories of each question, and used ChatGPT to annotate the primary subject based on the already annotated subject categories, then annotated secondary subjects based on the annotated primary subjects. This iterative annotation method can reduce the number of subjects input into ChatGPT, thereby reducing overhead.
Moreover, adopting this annotation approach has significantly saved the time spent on manually filtering Xiezhi-Speciality and Xiezhi-Interdiscipline. In the future, we plan to use the annotated data to further train the Annotation Model, aiming for its performance in subject annotation tasks to approach or surpass that of ChatGPT, thereby eliminating the need for ChatGPT for annotation.

\subsection{Bias, Ethical Problems and Social Impact}
\label{a5}
Even though we strive to avoid all possible ethical issues, we still find it challenging to guarantee the resolution of all ethical issues when it comes to constructing the largest knowledge-based evaluation benchmark in the world, with the largest number of discipline labels, largely compiled and translated from a single language.
In this section, we provide a detailed description of all the potential ethical issues that may exist in our dataset, along with how we alleviate them.

\textbf{Bias From Multilingual Dataset Curation} 

As a dataset constructed from Chinese sources, with the annotators being Chinese graduates, invariably, there is a level of Chinese bias.
These bias may come from a large number of Chinese related questions, Chinese Polical Standards or Chinese Distinct Value, which may result in a better Chinese understanding models will have better performance in Xiezhi.
We have made extensive efforts to eliminate such bias in Xiezhi.
Measures include: 
\begin{itemize} 
\item Selecting questions from Xiezhi-All that involve various countries and regions, rather than those biased towards Chinese contexts. 
\item Ensuring balanced distribution of virtual names in both Chinese and English styles, such as Li Hua in Chinese and Mike in English. 
\item Deleting questions involving political stances. 
\item Adding modified text like ``In China, ...'' to the beginning of questions involving common Chinese values.
\end{itemize} 

\textbf{Translation Version of Xiezhi} 

In this paper, we used Google Translate API to translate Xiezhi-Speciality and Xiezhi-Interdisciplinary into English, followed by extensive manual post-processing. 
The focus of verification includes several aspects: 
\begin{itemize} 
\item Correcting translation errors 
\item Correcting sentences with unnatural expressions 
\item Making precise expression for specific terms 
\end{itemize} 
We invited 10 annotators described in Appendix Manual Annotation to participate in the revision of the English version of Xiezhi. 
As all annotators are graduates who all deeply involved in writing several English papers, we are confident in their proficiency in both professional knowledge and English language use. 
Despite our high standards for the translated version, we still believe potential issues may exist in the current version of Xiezhi: 
\begin{itemize} 
\item Annotators are not native speakers living in English-speaking countries, so their expressions might not be perfectly idiomatic. 
\item The dataset covers 516 different fields. Although the annotators are graduate students with experience in English writing, precision in translating field-specific terms could be lacking. 
\end{itemize}

\textbf{Human Performance} 

We elaborated on the statistical method for human scoring in the experiment setting in Section Experiments. 
This scoring method compromises for the sheer reason that simply acquiring questions from one source would result in insufficient number and highly biased data. 
Moreover, as the content covered by each dataset varies greatly, we also found it hard to invite human participants from all 516 fields to provide human baselines.
We believe the human performance scores we provided to some extent reflect human performance on Xiezhi-Specialty and Xiezzi-Interdisciplinary.
However, the real-life decision-makers should not solely rely on these scores for these scores are only used as an comparison.

\textbf{Gender, Race, Religion, National Discrimination or Prejudice} 

The proposed of a benchmark will act as an indicator for LLM training for a period of time, if the benchmark itself harbors discriminatory or prejudiced content, it may encourage poor development of LLMs.
Therefore, in compliance with the NeurIPS dataset review standards, we eliminated all content related to gender, race, religion, and national discrimination or prejudice in the posterior process of Xiezhi-Specility and Xiezhi-Interdisciplinary as far as possible.
Our efforts in eliminating prejudiced content and discrimination are evident from the requirements listed in Table ~\ref{tab:example_in_annotation}.

Despite this, considering most of our questions were extracted from Chinese exam papers, and a small fraction generated from English papers, with annotation undertaken by Chinese postgraduate students, there still might be potential ethical issues in the Xiezhi: 
\begin{itemize} 
\item As our questions come from Chinese exams and English published papers, and our annotators are Chinese graduates, their annotation may unknowingly lean towards Eastern or Western cultural notions.
\item In consideration of the uneven distribution of gender, race, faith, and nationalities that can access the original questions, we resolved the issues of gender distribution in Xiezhi-Speciality and Xiezhi-Interdisciplinay. The implicit prejudice brought by race, faith, and nationality may be more severe, and though we ensured that our existing questions do not include prejudiced content in the question description and answers, we are unable to change the overall bias in the question distribution. 
\end{itemize}

The detail about how we follow the NeurIPS checklist and DataSheet is described in Appendix Checklist and Appendix Datasheet.

\subsection{Results on Other Metrics}
\label{a6}



Aside from the MRR score metric championed in our paper for ranking options, we have listed some other indicators to gauge the performance of different models on Xiezhi-Speciality and Xiezhi-Interdisciplinay. The variants brought about by the different indicators have also been analyzed.

We have also considered ranking in our metrics, employing Hit@1, Hit@4, and Mean Rank as indicators. The descriptions of these are as follows: We also utilized the conventional method of calculating accuracy, a method heavily employed in other papers.
\begin{itemize}
    \item Mean Rank~(MR): This measures the average rank position of a query concept's true parent among all candidates, divided by the total number of options.
    \item Hit@k: This is the number of query concepts whose parent is ranked in the top k positions, divided by the total number of queries.
    \item Accuracy: A standard measure used in most research.
\end{itemize}

Given the extensive computational cost, we carried out this experiment using only a subset of models. The results from different models, using different indicators, are presented in Tables ~\ref{tab:more_metric_sc}~\ref{tab:more_metric_dc}~\ref{tab:more_metric_se}~\ref{tab:more_metric_de}.

Our findings indicate that even when addressing the same dataset, different evaluation metrics yield different rankings. We suspect this may be because varying evaluation metrics unearth different characteristics encapsulated within the models. This is a significant factor in model evaluation and should be deeply investigated in order to draw comprehensive conclusions. Therefore, in our future work, we anticipate thoroughly researching the varying ranking results driven by different evaluation metrics.

\begin{table*}[t]
    \centering
    \resizebox{\textwidth}{!}{
    \myred{
    \begin{tabular}{|cc|cc|cc|cc|}
    \hline
        \multicolumn{8}{|c|}{Xiezhi-Speciality-English}\\
        \multicolumn{2}{|c}{MRR} & \multicolumn{2}{|c}{Hit@1} & \multicolumn{2}{|c}{Hit@4} & \multicolumn{2}{|c|}{MR$\downarrow$}  \\
        \hline
        GPT-4 & 0.476 & GPT-4 & 0.188 & GPT-4 & 0.641 & GPT-4 & 0.089 \\
        GPT-3.5 & 0.254 & GPT-3.5 & 0.101 & GPT-3.5 & 0.285 & GPT-3.5 & 0.204 \\
        \hline
         bloomz-mt & 0.386 & bloomz-mt & 0.096               & bloomz-mt & 0.268 
         & bloomz-7b1-mt & 0.317 \\
         bloomz-7b1 & 0.256 & bloomz-7b1 & 0.083              & bloomz-3b & 0.259 
         & bloomz-7b1 & 0.325 \\
         bloomz-7b1-mt & 0.242 & bloomz-7b1-mt & 0.080           & bloomz-7b1 & 0.250 
         & bloomz-mt & 0.333 \\
         falcon-7b & 0.223 & BELLE-7B-2M & 0.065             & bloomz-7b1-mt & 0.228 
         & llama-7b-hf & 0.339 \\
         baize-healthcare-lora-7B & 0.218 & bloomz-7b1-p3 & 0.061           & falcon-7b-instruct & 0.213 
         & falcon-40b-instruct & 0.394 \\
         bloomz-3b & 0.187 & baize-healthcare-lora-7B & 0.060& BELLE-7B-2M & 0.204 
         & falcon-7b-instruct & 0.424 \\
         bloomz-7b1-p3 & 0.173 & falcon-7b-instruct & 0.059      & falcon-7b & 0.186 
         & BELLE-7B-2M & 0.460 \\
         falcon-7b-instruct & 0.167 & bloomz-1b1 & 0.059              & bloomz-1b1 & 0.153 
         & falcon-7b & 0.462 \\
         bloomz-1b1 & 0.160 & bloomz-1b7 & 0.056              & bloomz-7b1-p3 & 0.150 
         & bloomz-3b & 0.490 \\
         bloomz-1b7 & 0.156 & falcon-40b-instruct & 0.055     & bloomz-1b7 & 0.146 
         & bloomz-1b1 & 0.494 \\
         \hline
         Random & 0.089 & Random & 0.020 & Random & 0.080 & Random & 0.510 \\
    \hline
    \end{tabular}}}
    \caption{
    \myred{
    The ranking of various models according to different metrics within the Xiezhi-Speciality-English Benchmark.
    }}
    \label{tab:more_metric_sc}
\end{table*}
\begin{table*}[t]
    \centering
    \resizebox{\textwidth}{!}{
    \myred{
    \begin{tabular}{|cc|cc|cc|cc|}
    \hline
        \multicolumn{8}{|c|}{Xiezhi-Interdiscipline-English}\\
        \multicolumn{2}{|c}{MRR} & \multicolumn{2}{|c}{Hit@1} & \multicolumn{2}{|c}{Hit@4} & \multicolumn{2}{|c|}{MR$\downarrow$}  \\
        \hline
        GPT-4 & 0.509 & GPT-4 & 0.243 & GPT-4 & 0.861 & GPT-4 & 0.071 \\
        GPT-3.5 & 0.382 & GPT-3.5 & 0.091 & GPT-3.5 & 0.371 & GPT-3.5 & 0.164 \\
        \hline
         bloomz-mt & 0.329 & bloomz-7b1 & 0.122              & bloomz-7b1 & 0.336                
         & bloomz-mt & 0.285 \\
         bloomz-7b1 & 0.203 & falcon-7b & 0.113               & bloomz-7b1-mt & 0.319             
         & bloomz-7b1 & 0.287 \\
         bloomz-7b1-mt & 0.195 & bloomz-7b1-mt & 0.107           & falcon-7b & 0.300                 
         & bloomz-7b1-mt & 0.370 \\
         baize-healthcare-lora-7B & 0.187 & baize-healthcare-lora-7B & 0.106& baize-healthcare-lora-7B & 0.267  
         & bloomz-3b & 0.412 \\
         BELLE-7B-2M & 0.161 & bloomz-mt & 0.086               & bloomz-mt & 0.210                 
         & llama-7b-hf & 0.428 \\
         falcon-7b-instruct & 0.159 & bloomz-1b1 & 0.078              & bloomz-1b7 & 0.171                
         & falcon-7b & 0.442 \\
         falcon-7b & 0.148 & bloomz-1b7 & 0.069              & bloomz-1b1 & 0.170                
         & falcon-7b-instruct & 0.428 \\
         bloomz-1b1 & 0.140 & bloomz-7b1-p3 & 0.062           & bloomz-560m & 0.145               
         & bloomz-7b1-p3 & 0.442 \\
         bloomz-7b1-p3 & 0.139 & bloomz-560m & 0.055             & llama-65b-hf & 0.130              
         & bloomz-1b1 & 0.475 \\
         bloomz-1b7 & 0.138 & falcon-7b-instruct & 0.053      & llama-30b-hf & 0.119              
         & bloomz-1b7 & 0.476 \\
         \hline
         Random & 0.089 & Random & 0.020 & Random & 0.080 & Random & 0.510 \\
    \hline
    \end{tabular}}}
    \caption{
    \myred{
    The ranking of various models according to different metrics within the Xiezhi-Interdiscipline-English Benchmark.
    }}
    \label{tab:more_metric_dc}
\end{table*}\begin{table*}[t]
    \centering
    \resizebox{\textwidth}{!}{
    \myred{
    \begin{tabular}{|cc|cc|cc|cc|}
    \hline
        \multicolumn{8}{|c|}{Xiezhi-Speciality-Chinese}\\
        \multicolumn{2}{|c}{MRR} & \multicolumn{2}{|c}{Hit@1} & \multicolumn{2}{|c}{Hit@4} & \multicolumn{2}{|c|}{MR$\downarrow$}  \\
        \hline
        GPT-4 & 0.371 & GPT-4 & 0.104 & GPT-4 & 0.537 & GPT-4 & 0.197 \\
        GPT-3.5 & 0.243 & GPT-3.5 & 0.064 & GPT-3.5 & 0.376 & GPT-3.5 & 0.261 \\
        \hline
         bloomz-mt & 0.324 & BELLE-7B-0.6M & 0.089           & baize-lora-7B & 0.304             
         & llama-65b-hf & 0.344 \\
         baize-lora-7B & 0.198 & baize-lora-7B & 0.087           & BELLE-7B-0.6M & 0.260             
         & llama-7b-hf & 0.354 \\
         BELLE-7B-0.6M & 0.190 & vicuna-7b-delta-v1.1 & 0.083    & vicuna-13b-delta-v1.1 & 0.257     
         & bloomz-3b & 0.387 \\
         bloomz-3b & 0.188 & bloomz-7b1-p3 & 0.082           & bloomz-3b & 0.256                 
         & BELLE-7B-0.6M & 0.410 \\
         bloomz-7b1-p3 & 0.187 & vicuna-13b-delta-v1.1 & 0.081   & bloomz-mt & 0.241                    
         & bloomz-7b1-p3 & 0.419 \\
         vicuna-13b-delta-v1.1 & 0.184 & bloomz-mt & 0.076                  & vicuna-7b-delta-v1.1 & 0.238      
         & gpt-neox-20b & 0.426 \\
         vicuna-7b-delta-v1.1 & 0.181 & BELLE-7B-0.2M & 0.075           & bloomz-1b7 & 0.235                
         & dolly-v2-3b & 0.427 \\
         BELLE-7B-0.2M & 0.178 & BELLE-7B-2M & 0.074             & baize-healthcare-lora-7B & 0.233  
         & bloomz-mt & 0.430 \\
         baize-healthcare-lora-7B & 0.177 & BELLE-7B-1M & 0.072             & bloomz-7b1-p3 & 0.232             
         & llama-13b-hf & 0.439 \\
         BELLE-7B-1M & 0.176 & baize-healthcare-lora-7B & 0.071& BELLE-7B-1M & 0.231               
         & pythia-1.4b & 0.450 \\
         \hline
         Random & 0.089 & Random & 0.020 & Random & 0.080 & Random & 0.510 \\
    \hline
    \end{tabular}}}
    \caption{
    \myred{
    The ranking of various models according to different metrics within the Xiezhi-Speciality-Chinese Benchmark.
    }}
    \label{tab:more_metric_se}
\end{table*}\begin{table*}[t]
    \centering
    \resizebox{\textwidth}{!}{
    \myred{
    \begin{tabular}{|cc|cc|cc|cc|}
    \hline
        \multicolumn{8}{|c|}{Xiezhi-Interdiscipline-Chinese}\\
        \multicolumn{2}{|c}{MRR} & \multicolumn{2}{|c}{Hit@1} & \multicolumn{2}{|c}{Hit@4} & \multicolumn{2}{|c|}{MR$\downarrow$}  \\
        \hline
        GPT-4 & 0.413 & GPT-4 & 0.161 & GPT-4 & 0.714 & GPT-4 & 0.103 \\
        GPT-3.5 & 0.214 & GPT-3.5 & 0.084 & GPT-3.5 & 0.237 & GPT-3.5 & 0.188 \\
        \hline
         baize-lora-13B & 0.209 & baize-lora-13B & 0.108              & BELLE-7B-0.6M & 0.269             
         & BLLE-7B-0.6M & 0.384 \\
         BELLE-7B-0.6M & 0.198 & bloomz-mt & 0.100    & baize-lora-13B & 0.252            
         & llama-13b-hf & 0.396 \\
         alpaca-lora-7b & 0.192 & bloomz-7b1-p3 & 0.095               & alpaca-lora-7b & 0.247            
         & bloomz-7b1-p3 & 0.405 \\
         bloomz-7b1-p3 & 0.188 & bloomz-7b1 & 0.083                  & BELLE-7B-1M & 0.244               
         & stablelm-tuned-alpha-7b & 0.419 \\
         baize-lora-7B & 0.186 & BELLE-7B-0.6M & 0.081               & baize-lora-7B & 0.234             
         & bloomz-mt & 0.427 \\
         bloomz-mt & 0.185 & BELLE-7B-1M & 0.080                 & BELLE-7B-0.2M & 0.229             
         & falcon-40b-instruct & 0.435 \\
         BELLE-7B-1M & 0.180 & BELLE-7B-0.2M & 0.077               & bloomz-mt & 0.220  
         & llama-30b-hf & 0.441 \\
         BELLE-7B-0.2M & 0.173 & bloomz-7b1-mt & 0.076               & bloomz-7b1-p3 & 0.207             
         & pythia-12b & 0.455 \\
         bloomz-7b1 & 0.166 & baize-lora-30B & 0.075              & BELLE-7B-2M & 0.204              
         & BELLE-7B-1M & 0.465 \\
         BELLE-7B-2M & 0.164 & bloomz-3b & 0.075                   & falcon-40b-instruct & 0.200       
         & dolly-v2-7b & 0.465 \\
         \hline
         Random & 0.089 & Random & 0.020 & Random & 0.080 & Random & 0.510 \\
    \hline
    \end{tabular}}}
    \caption{
    \myred{
    The ranking of various models according to different metrics within the Xiezhi-Interdiscipline-Chinese Benchmark.
    }}
    \label{tab:more_metric_de}
\end{table*}

}

\subsection{Models}
\label{a7}

A comprehensive overview of the evaluated models is presented in Table~\ref{tab:models}. 
The ``Model'' column specifies the names of the analyzed models, while the ``\#Parameter'' column indicates their respective parameters. 
The ``Base Model'' column reveals the origins of the fine-tuned models and a dash~(-) signifies that it is not an instruction fine-tuned model. 
The number of Transformer layers utilized in each model is denoted by the ``\#Layer'' column, and the individual encoder and decoder Transformer layers are indicated by the ``\#Encoder'' and ``\#Decoder'' columns, respectively.
Lastly, the ``\#IFT Sample'' column represents the quantity of instruction samples employed for instruction fine-tuning.


\begin{table*}
    \small
    \centering
    \resizebox{1\textwidth}{!}{
    \begin{tabular}{c|c|c|c|c|c|c|c}
        \hline
        \textbf{Model}& \textbf{\#Parameter} & \textbf{Base Model} & \textbf{\#Layer} & \textbf{\#Encoder} & \textbf{\#Decoder} & \textbf{\#Pretrain Tokens} & \textbf{\#IFT Sample} \\
        \hline
        BLOOM-560m~\cite{scao2022bloom}~\footnote{https://huggingface.co/bigscience/bloom-560m} & 0.56B & - & 24 & - & 24 & 350B tokens & -  \\
        BLOOMZ-560m~\cite{muennighoff2022crosslingual}~\footnote{https://huggingface.co/bigscience/bloomz-560m} & 0.56B & BLOOM-560m & 24 & - & 24 & - & 3.67B tokens \\
        Pythia-1B~\cite{biderman2023pythia}~\footnote{https://huggingface.co/EleutherAI/pythia-1b} & 1B & - & 16 & - & 16 & 300B tokens & - \\
        BLOOM-1b7~\cite{scao2022bloom}~\footnote{https://huggingface.co/bigscience/bloom-1b7} & 1.7B & - & 24 & - & 24 & 350B tokens & - \\
        BLOOMZ-1b7~\cite{muennighoff2022crosslingual}~\footnote{https://huggingface.co/bigscience/bloomz-1b7} & 1.7B & BLOOM-1b7 & 24 & - & 24 & - & 8.39B tokens \\
        Dolly-v2-3b~\cite{2023dolly}~\footnote{https://huggingface.co/databricks/dolly-v2-3b} & 2.8B & Pythia-2.8B & 32 & - & 32 & - & 15K \\
        Pythia-2.8B~\cite{biderman2023pythia}~\footnote{https://huggingface.co/EleutherAI/pythia-2.8b} & 2.8B & - & 32 & - & 32 & 300B tokens & - \\
        BLOOM-3b~\cite{scao2022bloom}~\footnote{https://huggingface.co/bigscience/bloom-3b} & 3B & - & 30 & - & 30 & 350B tokens & - \\
        BLOOMZ-3b~\cite{muennighoff2022crosslingual}~\footnote{https://huggingface.co/bigscience/bloomz-3b} & 3B & BLOOM-3b & 30 & - & 30 & - & 8.39B tokens \\
        StableLM-Tuned-Alpha-3B~\cite{2023StableLM}~\footnote{https://huggingface.co/stabilityai/stablelm-tuned-alpha-3b} & 3B & StableLM-Base-Alpha-3B & 16 & - & 16 & - & 632K \\
        ChatGLM-6B~\cite{zeng2023glm-130b,du2022glm}~\footnote{https://huggingface.co/THUDM/chatglm-6b} & 6B & - & 28 & 28 & 28 & 1T tokens & \checkmark \\
        DoctorGLM~\cite{xiong2023doctorglm}~\footnote{https://github.com/xionghonglin/DoctorGLM} & 6B & ChatGLM-6B & 28 & 28 & 28 & - & 6.38M \\
        Dolly-v2-7b~\cite{2023dolly}~\footnote{https://huggingface.co/databricks/dolly-v2-7b} & 6.9B & Pythia-6.9B & 32 & - & 32 & - & 15K \\
        h2ogpt-oig-oasst1-512-6.9b~\cite{2023h2ogpt}~\footnote{https://huggingface.co/h2oai/h2ogpt-oig-oasst1-512-6.9b} & 6.9B & Pythia-6.9B & 32 & - & 32 & - & 398K \\
        Pythia-6.9B~\cite{biderman2023pythia}~\footnote{https://huggingface.co/EleutherAI/pythia-6.9b} & 6.9B & - & 32 & - & 32 & 300B tokens & - \\
        Alpaca-7B~\cite{alpaca}~\footnote{https://huggingface.co/tatsu-lab/alpaca-7b-wdiff} & 7B & LLaMA-7B & 32 & - & 32 & - & 52K \\
        Alpaca-LoRA-7B~\cite{2023alpacalora}~\footnote{https://huggingface.co/tloen/alpaca-lora-7b} & 7B & LLaMA-7B & 32 & - & 32 & - & 52K \\
        Baize-7B~\cite{xu2023baize}~\footnote{https://huggingface.co/project-baize/baize-lora-7B} & 7B & LLaMA-7B & 32 & - & 32 & - & 263K \\
        Baize Healthcare-7B~\cite{xu2023baize}~\footnote{https://huggingface.co/project-baize/baize-healthcare-lora-7B} & 7B & LLaMA-7B & 32 & - & 32 & - & 201K \\
        LLaMA-7B~\cite{touvron2023llama}~\footnote{https://huggingface.co/decapoda-research/llama-7b-hf} & 7B & - & 32 & - & 32 & 1T tokens & - \\
        StableLM-Tuned-Alpha-7B~\cite{2023StableLM}~\footnote{https://huggingface.co/stabilityai/stablelm-tuned-alpha-7b} & 7B & StableLM-Base-Alpha-7B & 16 & - & 16 & - & 632K \\
        Vicuna-7b-delta-v1.1~\cite{vicuna2023}~\footnote{https://github.com/lm-sys/FastChat\#vicuna-weights} & 7B & LLaMA-7B & 32 & - & 32 & - & 70K \\
        BELLE-7B-0.2M~\footnote{https://huggingface.co/BelleGroup/BELLE-7B-0.2M}/0.6M~\footnote{https://huggingface.co/BelleGroup/BELLE-7B-0.6M}/1M~\footnote{https://huggingface.co/BelleGroup/BELLE-7B-1M}/2M~\footnote{https://huggingface.co/BelleGroup/BELLE-7B-2M}~\cite{belle2023exploring} & 7.1B & Bloomz-7b1-mt & 30 & - & 30 & - & 0.2M/0.6M/1M/2M \\
        BLOOM-7b1~\cite{scao2022bloom}~\footnote{https://huggingface.co/bigscience/bloom-7b1} & 7.1B & - & 30 & - & 30 & 350B tokens & - \\
        BLOOMZ-7b1~\footnote{https://huggingface.co/bigscience/bloomz-7b1}/mt~\footnote{https://huggingface.co/bigscience/bloomz-7b1-mt}/p3~\footnote{https://huggingface.co/bigscience/bloomz-7b1-p3}~\cite{muennighoff2022crosslingual} & 7.1B & BLOOM-7b1 & 30 & - & 30 & - & 4.19B tokens \\
        Dolly-v2-12b~\cite{2023dolly}~\footnote{https://huggingface.co/databricks/dolly-v2-12b} & 12B & Pythia-12B & 36 & - & 36 & - & 15K \\
        h2ogpt-oasst1-512-12b~\cite{2023h2ogpt}~\footnote{https://huggingface.co/h2oai/h2ogpt-oasst1-512-12b} & 12B & Pythia-12B & 36 & - & 36 & - & 94.6K \\
        Open-Assistant-SFT-4-12B~\cite{2023openassistant}~\footnote{https://huggingface.co/OpenAssistant/oasst-sft-4-pythia-12b-epoch-3.5} & 12B & Pythia-12B-deduped & 36 & - & 36 & - & 161K \\
        Pythia-12B~\cite{biderman2023pythia}~\footnote{https://huggingface.co/EleutherAI/pythia-12b} & 12B & - & 36 & - & 36 & 300B tokens & - \\
        Baize-13B~\cite{xu2023baize}~\footnote{https://huggingface.co/project-baize/baize-lora-13B} & 13B & LLaMA-13B & 40 & - & 40 & - & 263K \\
        LLaMA-13B~\cite{touvron2023llama}~\footnote{https://huggingface.co/decapoda-research/llama-13b-hf} & 13B & - & 40 & - & 40 & 1T tokens & - \\
        Vicuna-13b-delta-v1.1~\cite{vicuna2023}~\footnote{https://github.com/lm-sys/FastChat\#vicuna-weights} & 13B & LLaMA-13B & 40 & - & 40 & - & 70K \\
        moss-moon-003-sft~\cite{2023moss}~\footnote{https://huggingface.co/fnlp/moss-moon-003-sft} & 16B & moss-moon-003-base & 34 & - & 34 & - & 1.1M \\
        moss-moon-003-sft-plugin~\cite{2023moss}~\footnote{https://huggingface.co/fnlp/moss-moon-003-sft-plugin} & 16B & moss-moon-003-base & 34 & - & 34 & - & 1.4M \\
        GPT-NeoX-20B~\cite{gptneox}~\footnote{https://huggingface.co/EleutherAI/gpt-neox-20b} & 20B & - & 44 & - & 44 & 825GB & - \\
        h2ogpt-oasst1-512-20b~\cite{2023h2ogpt}~\footnote{https://huggingface.co/h2oai/h2ogpt-oasst1-512-20b} & 20B & GPT-NeoX-20B & 44 & - & 44 & - & 94.6K \\
        Baize-30B~\cite{xu2023baize}~\footnote{https://huggingface.co/project-baize/baize-lora-30B} & 33B & LLaMA-30B & 60 & - & 60 & - & 263K \\
        LLaMA-30B~\cite{touvron2023llama}~\footnote{https://huggingface.co/decapoda-research/llama-30b-hf} & 33B & - & 60 & - & 60 & 1.4T tokens & - \\
        LLaMA-65B~\cite{touvron2023llama}~\footnote{https://huggingface.co/decapoda-research/llama-65b-hf} & 65B & - & 80 & - & 80 & 1.4T tokens & - \\
        BLOOM~\cite{scao2022bloom}~\footnote{https://huggingface.co/bigscience/bloom} & 176B & - & 70 & - & 70 & 366B tokens & - \\
        BLOOMZ~\footnote{https://huggingface.co/bigscience/bloomz}/mt~\footnote{https://huggingface.co/bigscience/bloomz-mt}/p3~\footnote{https://huggingface.co/bigscience/bloomz-p3}~\cite{muennighoff2022crosslingual} & 176B & BLOOM & 70 & - & 70 & - & 2.09B tokens \\
        ChatGPT~(2023.05.01)~\cite{openaichatgpt}~\footnote{https://platform.openai.com/docs/models/gpt-3-5} & - & GPT-3.5 & - & - & - & - & \checkmark \\
        GPT-4~(2023.05.01)~\cite{openai2023gpt4}~\footnote{https://platform.openai.com/docs/models/gpt-4} & - & - & - & - & - & - & \checkmark \\
        \hline
    \end{tabular}
    }
    \vspace{3mm}
    \caption{The models we evaluted in our experiments.}
    \label{tab:models}
\end{table*}




\subsection{Data Sheet}
\label{sec:datasheet}
\myred{
\subsection{Motivation}

\textbf{For what purpose was the dataset created? Was there a specific task in mind? Was there a specific gap that needed to be filled? Please provide a description}

Xiezhi was created for the purpose of comprehensively evaluating the domain knowledge capabilities of large language models~(LLMs).
Some key gaps and needs that existing benchmarks did not adequately address:
\begin{enumerate}
    \item Existing benchmarks did not cover enough tasks or domains to fully assess the breadth of knowledge and capabilities of advanced LLMs.
    \item Many existing benchmarks quickly became outdated as they got incorporated into the training data of the latest LLMs. There was a need for benchmarks with fresher data.
    \item Most benchmarks relied on 4-option multiple choice questions. This made it too easy for models to guess correctly. More options were needed to better differentiate model capabilities.
    \item Existing evaluation methods using multiple choice extractions had limitations for generative models. A better evaluation approaches are needed and Xiezhi propose to rank options by generative probability.
\end{enumerate}

\textbf{Who created this dataset (e.g., which team, research group) and on behalf of which entity (e.g., company, institution, organization)?}

The Knowledge Works Research Laboratory from Fudan University in China created this dataset.

\textbf{Who funded the creation of the dataset? If there is an associated grant, please provide the name of the grantor and the grant name and number.}

The grant come from Fudan University.

\subsection{Composition}

\textbf{What do the instances that comprise the dataset represent (e.g., documents, photos, people, countries)? Are there multiple types of instances (e.g., movies, users, and ratings; people and interactions between them; nodes and edges)? Please provide a description.}

The instances that comprise the Xiezhi are multiple choice questions designed to assess domain knowledge across a wide range of disciplines with the following components:
\begin{itemize}

    \item Question text: The question or problem statement.
    \item Answer options: 4 possible options to choose from, with 1 correct answer and 3 near misses.
    \item Correct answer: The ground truth answer out of the 4 options.
    \item Subject labels: One or more labels categorizing the discipline/domain of knowledge required to answer the question correctly (516 total subjects organized hierarchically into 13 top-level categories).

\end{itemize}

\textbf{How many instances are there in total (of each type, if appropriate)?}

Xiezhi-All consist of 249,587 questions, Xiezhi-Speciality consists of 14,041 questions, Xiezhi-Interdiscipline consists of 10,746 questions, Xiezhi-Meta consists of 20,124 questions and Xiezhi-Trian consists of 2,555 questions.

\textbf{Does the dataset contain all possible instances or is it a sample (not necessarily random) of instances from a larger set? If the dataset is a sample, then what is the larger set? Is the sample representative of the larger set (e.g., geographic coverage)? If so, please describe how this representativeness was validated/verified. If it is not representative of the larger set, please describe why not (e.g., to cover a more diverse range of instances, because instances were withheld or unavailable).}

Xiezhi-All contain all possible instances, but we don't tend to open source it for it is not verfied by human.
Xiezhi-Meta, Xiezhi-Train, Xiezhi-Speciality and Xiezhi-Interdiscipline are subset of Xiezhi-All but undertook manual verification. 

\textbf{What data does each instance consist of? ``Raw'' data (e.g., unprocessed text or images)or features? In either case, please provide a description.}

Multiple-choice questions with manual semantic annotations. Please refer to Fig.~\ref{fig:questions} for more details.

\textbf{Is any information missing from individual instances? If so, please provide a description, explaining why this information is missing (e.g., because it was unavailable). This does not include intentionally removed information, but might include, e.g., redacted text.}

Individual instances within the Xiezhi-Meta, Xiezhi-Train, Xiezhi-Speciality, and Xiezhi-Interdiscipline datasets have undergone manual verification, resulting in a complete data set with no instances missing information. However, Xiezhi-All, containing all raw data from open-source exams, has a limited amount of data missing since it has been annotated only through automatic annotation models.

\textbf{Are relationships between individual instances made explicit (e.g., users' movie ratings, social network links)? If so, please describe how these relationships are made explicit.
}

Individual instances are extracted from all kinds of Chinese Examinations, so these instances may share the same level of difficulty or discipline labels to some extent.

\textbf{Are there recommended data splits (e.g., training, development/validation, testing)?}

Yes, we propose Xiezhi-Train of experiments in demonstration setting, and Xiezhi-Meta for model training, we also propose Xiezhi-Interdiscipline and Xiezhi-Specility for model testing.

\textbf{Are there any errors, sources of noise, or redundancies in the dataset? If so, please provide a description.}

The dataset is carefully reviewed and checked automatically and manually with a strict quality control protocol, so there will be few error or noise.

\textbf{Is the dataset self-contained, or does it link to or otherwise rely on external resources (e.g.,websites, tweets, other datasets)?}

The dataset is self-contained.

\textbf{Does the dataset contain data that might be considered confidential (e.g., data that is protected by legal privilege or by doctor– patient confidentiality, data that includes the content of individuals’ non-public communications)? If so, please provide a description.}
No.

\textbf{Does the dataset contain data that, if viewed directly, might be offensive, insulting, threatening, or might otherwise cause anxiety? If so, please describe why}

The dataset is carefully reviewed, all data might be offensive, insulting, threatening or might otherwise cause anxiety are excluded automatically and manually.

\textbf{Does the dataset identify any subpopulations (e.g., by age, gender)?}

No.

\textbf{Is it possible to identify individuals (i.e., one or more natural persons), either directly or indirectly (i.e., in combination with other data) from the dataset?}

No, all the instances in Xiezhi are questions about domain knowledge, it is impossible to identify individuals from the dataset.

\textbf{Does the dataset contain data that might be considered sensitive in any way (e.g., data that reveals race or ethnic origins, sexual orientations, religious beliefs, political opinions or union memberships, or locations; financial or health data; biometric or genetic data; forms of government identification, such as social security numbers; criminal history)?}

No.

\subsection{Collection Process}
\textbf{How was the data associated with each instance acquired? Was the data directly observable (e.g., raw text, movie ratings), reported by subjects (e.g., survey responses), or indirectly inferred/derived from other data (e.g., part-of-speech tags, model-based guesses for age or language)? If data was reported by subjects or indirectly inferred/derived from other data, was the data validated/verified? If so, please describe how.}

All the questions are extracted from all kinds of Chinese examinations online or generated from Chinese academic surveys, so the answers, options and questions are all directly observable.

\textbf{What mechanisms or procedures were used to collect the data (e.g., hardware apparatus or sensor, manual human curation, software program, software API)? How were these mechanisms or procedures validated?}

We used a Python crawler to grab questions from an online site and used ChatGPT to automatically annotate the questions.

\textbf{Who was involved in the data collection has process (e.g., students, crowdworkers, contractors) and how were they compensated (e.g., how much were crowdworkers paid)?}

The questions are collected and generated by the authors, and are manual annoated by 10 Chinese graduates.
All the annotators are paid above the local minimum wage.

\textbf{Over what timeframe was the data collected? Does this timeframe match the creation time frame of the data associated with the instances (e.g., recent crawl of old news articles)? If not, please describe the timeframe in which the data associated with the instances was created.}

The data was collected from March 2023 until May 2023, it match the created time of our github.

\textbf{Were any ethical review processes conducted (e.g., by an institutional review board)?}

We undertake a serious ethical review, please refer to Appendix Bias, Ethical Problems and Social Impact for more details.

\textbf{Did you collect the data from the individuals in question directly, or obtain it via third parties or other sources (e.g., websites)?}

We do not collect the data from any individuals.

\textbf{Did the individuals in question consent to the collection and use of their data?}

N/A

\textbf{If consent was obtained, were the consenting individuals provided with a mechanism to revoke their consent in the future or for certain uses?}

N/A

\textbf{Has an analysis of the potential impact of the dataset and its use on data subjects (e.g., a data protection impact analysis) been conducted?}

N/A

\subsection{Preprocessing / Cleaning / Labeling}

\textbf{Was any preprocessing/cleaning/labeling of the data done (e.g., discretization or bucketing, tokenization, part-of-speech tagging, SIFT feature extraction, removal of instances, processing of missing values)?}

Yes, our preprocessing process is illustrated in Fig.~\ref{fig:framework}.

\textbf{Was the ``raw'' data saved in addition to the preprocessed/cleaned/labeled data (e.g., to support unanticipated future uses)? If so, please provide a link or other access point to the ``raw'' data}

Xiezhi-All saved all the raw data, the github url will be released after the reviewing of AAAI-2024.

\textbf{Is the software used to preprocess/clean/label the instances available? If so, please provide a link or other access point.}

No.

\subsection{Uses}

\textbf{Has the dataset been used for any tasks already?}

No, not yet.

\textbf{Is there a repository that links to any or all papers or systems that use the dataset?}

No.

\textbf{What (other) tasks could the dataset be used for?}

The dataset could be used for instruction-tuning for boosting LLMs performance in domain text understanding.

\textbf{Is there anything about the composition of the dataset or the way it was collected and preprocessed/cleaned/labeled that might impact future uses? For example, is there anything that a dataset consumer might need to know to avoid uses that could result in unfair treatment of individuals or groups (e.g., stereotyping, quality of service issues) or other risks or harms (e.g., legal risks, financial harms)? If so, please provide a description. Is there anything a dataset consumer could do to mitigate these risks or harms?}

No.

\textbf{Are there tasks for which the dataset should not be used? If so, please provide a description.}

No.

\subsection{Distribution}

\textbf{Will the dataset be distributed to third parties outside of the entity (e.g., company, institution, organization) on behalf of which the dataset was created?}

Yes, probably.

\textbf{How will the dataset will be distributed (e.g., tarball on website, API, GitHub)? Does the dataset have a digital object identififier (DOI)?}

The dataset will be distributed at \url{https://github.com/MikeGu721/XiezhiBenchmark}

\textbf{When will the dataset be distributed?}

The up-to-date dataset has been uploaded now.

\textbf{Will the dataset be distributed under a copyright or other intellectual property (IP) license, and/or under applicable terms of use (ToU)? If so, please describe this license and/or ToU, and provide a link or other access point to, or otherwise reproduce, any relevant licensing terms or ToU, as well as any fees associated with these restrictions.}

This dataset is released under the CC BY-SA 4.0 license for general research purposes.

\textbf{Have any third parties imposed IP-based or other restrictions on the data associated with the instances? If so, please describe these restrictions, and provide a link or other access point to, or otherwise reproduce, any relevant licensing terms, as well as any fees associated with these restrictions.}

No.

\textbf{Do any export controls or other regulatory restrictions apply to the dataset or to individual instances? If so, please describe these restrictions, and provide a link or other access point to, or otherwise reproduce, any supporting documentation.}

No.

\subsection{Maintance}

\textbf{Who is supporting / hosting / maintaining the dataset?}

Cognitive Understanding Group of Knowledge Works Research Laboratory from Fudan University, China.

\textbf{How can the owner / curator / manager of the dataset be contacted (e.g., email address)?}

The emails of the first authors are \{zhgu22, xxzhu22\}@m.fudan.edu.cn, and the corrsponding authors are \{hwfeng, shawyh\}@fudan.edu.cn.

\textbf{Is there an erratum?}

No.

\textbf{Will the dataset be updated (e.g., to correct labeling errors, add new instances, delete instances)?}

According to our current plans, the dataset will be updated twice a year. 

\textbf{If the dataset relates to people, are there applicable limits on the retention of the data associated with the instances (e.g., were the individuals in question told that their data would be retained for a fixed period of time and then deleted)?}

N/A.

\textbf{Will older versions of the dataset continue to be supported / hosted / maintained?}

Yes, older version is still maintained and updated and will be communicated to users via Github.

\textbf{If others want to extend/augment/build on/contribute to the dataset, is there a mechanism for them to do so? }

We welcome people from all walks of life to use our data, and as we mentioned in Sec Discussion, the rapid development of big models requires more challenging datasets, more evaluation metrics and evaluation methods, and we are more than willing to contribute Xiezhi to this great goal.
}
\subsection{Check List}
\label{sec:checklist}
\myred{
\begin{enumerate}
    
\item  For all authors...
\begin{enumerate}
    \item Do the main claims made in the abstract and introduction accurately reflect the paper’s contributions and scope? 
    
    \blue{[Yes]}
    
    \item Did you describe the limitations of your work? 
    
    \blue{[Yes]} We describe the limitations in Section Discuss.
    
    \item Did you discuss any potential negative societal impacts of your work?
    
    \blue{[Yes]} We describe the limitations in Appendix Bias, Ethical Problems and Social Impact.
    
    \item Have you read the ethics review guidelines and ensured that your paper conforms to them?
    
    \blue{[Yes]}
\end{enumerate}

\item If you are including theoretical results...

\begin{enumerate}
\item Did you state the full set of assumptions of all theoretical results? 

\blue{[N/A]}

\item Did you include complete proofs of all theoretical results? 

\blue{[N/A]}

\end{enumerate}

\item If you ran experiments (e.g. for benchmarks)...

\begin{enumerate}
\item Did you include the code, data, and instructions needed to reproduce the main experimental results (either in the supplemental material or as a URL)? 

\blue{[Yes]} We describe our experiment setting in Section Experiments and the random seed in Appendix Detail Hyper-parameters.

\item Did you specify all the training details (e.g., data splits, hyperparameters, how they
were chosen)? 

\blue{[Yes]} We describe the training of proposed annotation in Appendix Auto Annotator.

\item Did you report error bars (e.g., with respect to the random seed after running experiments multiple times)? 

\blue{[No]} The experiments consume substantial GPU resources; therefore, to ensure consistency, all our experiments were conducted under the random seed of 42.

\item Did you include the total amount of compute and the type of resources used (e.g., type
of GPUs, internal cluster, or cloud provider)? 

\blue{[Yes]} Provided in Appendix Detail Hyper-parameters.

\end{enumerate}

\item If you are using existing assets (e.g., code, data, models) or curating/releasing new assets...

\begin{enumerate}

\item If your work uses existing assets, did you cite the creators? 

\blue{[Yes]} We use baseline models from Huggingface's Transformers, and detail describe and cite them in Appendix Models.

\item Did you mention the license of the assets? 

\blue{[Yes]}

\item Did you include any new assets either in the supplemental material or as a URL? 

\blue{[Yes]} We provide details in Appendix Models.

\item Did you discuss whether and how consent was obtained from people whose data you’re
using/curating? 

\blue{[N/A]}

\item Did you discuss whether the data you are using/curating contains personally identifiable information or offensive content? 

\blue{[N/A]}

\end{enumerate}

\item If you used crowdsourcing or conducted research with human subjects...

\begin{enumerate}

\item Did you include the full text of instructions given to participants and screenshots, if applicable? 

\blue{[Yes]} The instructions are included in Appendix Manual Annotation.

\item Did you describe any potential participant risks, with links to Institutional Review
Board (IRB) approvals, if applicable? 

\blue{[N/A]}

\item Did you include the estimated hourly wage paid to participants and the total amount
spent on participant compensation? 

\blue{[Yes]}
\end{enumerate}

\end{enumerate}
}

\end{CJK}
\end{document}